\documentclass[envcountsame,leqno]{llncs}

\usepackage[T1]{fontenc}
\usepackage{pstricks,pst-node,pst-tree}
\usepackage{amsmath, amsfonts, amssymb}
\usepackage[sans]{dsfont}
\usepackage{yfonts}
\usepackage{bbm}
\usepackage{graphicx}

\newcommand{\U}{\mathcal{U}}
\newcommand{\IV}{induction hypothesis\;}

\newcommand{\csfw}[3]{\mbox{$(#1  \clsr  #2) \in #3$}}

\newcommand{\tobedef}{\circ_R}
\newcommand{\beq}{\begin {quote}}
\newcommand{\enq}{\end{quote}}
\newcommand{\mun}{\sqcup}
\newcommand{\mint}{\sqcap}

\newcommand{\longreport}[2]{#2}
\newcommand{\ignore}[1]{}
\newcommand{\nmr}[2]{\begin{array}{l}#1\vspace{0.#2cm}\end{array}}

\newcommand{\CSL}{\mathcal{CSL}}

\newcommand{\I}{\mathcal{I}}
\newcommand{\J}{\mathcal{J}}

\newcommand{\Vp}{\mathcal{V}_{p}}

\newcommand{\M}{\mathcal{M}}

\newcommand{\clsr}{\leftleftarrows}
\newcommand{\impl}{\rightarrow}
\newcommand{\Lcsl}{\mathcal{L}_{\CSL}}
\newcommand{\CSMS}{\mathbf{CSMS}}

\newcommand{\Mc}{{\mathcal{M}_C}}

\newcommand{\sqset}{\sqsubseteq}
\newcommand{\csim}{\clsr}

\spnewtheorem{claim2}{Claim}{\bfseries}{\itshape}
\spnewtheorem{step}{Step}{\bfseries}{\normalfont}

\title{Comparative concept similarity over Minspaces: Axiomatisation and Tableaux Calculus}

\author{R\'{e}gis Alenda\inst{1} \and Nicola Olivetti\inst{1} \and Camilla Schwind\inst{2}}

\institute{
LSIS - UMR CNRS 6168  \\
Domaine Universitaire de Saint-J\'{e}r\^{o}me, Avenue Escadrille Normandie-Niemen ,\\
13397 MARSEILLE CEDEX 20\\
\texttt{regis.alenda@lsis.org} et
\texttt{nicola.olivetti@univ-cezanne.fr}
\and
LIF - UMR CNRS 6166\\
Centre de Math\'{e}matiques et Informatique \\
39 rue Joliot-Curie - F-13453 Marseille Cedex13.  \\
\texttt{camilla.schwind@lif.univ-mrs.fr} 
}

\begin{document}

\maketitle

\begin{abstract}
We study the logic of comparative concept similarity $\CSL$ introduced by  Sheremet, Tishkovsky, Wolter and Zakharyaschev to capture a form of  qualitative  similarity comparison.  In this logic we can  formulate assertions of the form " objects A are more similar to B than to C". The semantics of this logic is defined by structures equipped with distance functions evaluating the similarity degree of objects.  We consider here the particular case of the semantics induced by \emph{minspaces},  the latter being  distance spaces where the minimum of a set of distances always exists. It turns out that the semantics over arbitrary minspaces can be equivalently specified  in terms of preferential structures, typical of conditional logics. We first give  a direct axiomatisation of this logic over Minspaces.  We next define a decision procedure in the form of  a tableaux calculus.  Both the  calculus and the axiomatisation take advantage of  the reformulation of the semantics in terms of preferential structures.

\end{abstract}

\section{Introduction}\label{section:introduction}
The logics of comparative concept similarity $\CSL$ have been  proposed by
Sheremet, Tishkovsky, Wolter et Zakharyaschev in \cite{Wolter:LPAR05} to capture a form of
qualitative comparison between concept instances. In these logics we can
express assertions or judgments  of the form: "Renault Clio is more similar
to Peugeot 207 than to WW Golf". These logics may  find an  application in ontology
languages, whose logical base is provided by Description Logics (DL), allowing concept definitions based on proximity/similarity measures.
For instance \cite{Wolter:LPAR05}, the color "Reddish " may  be defined as a color which is more similar to a prototypical "`Red"' than to any other color (in some color model as RGB). The aim is to dispose of a language  in which logical classification provided by standard DL is  integrated  with   classification mechanisms based on  calculation of proximity measures. The latter is typical for instance of domains like bio-informatics or linguistics.
In a series of papers \cite{Wolter:LPAR05,Wolter08,conf/birthday/KuruczWZ05,DBLP:journals/logcom/SheremetTWZ07} the authors proposes several languages comprising absolute similarity measures and  comparative similarity operator(s). 
In this paper we consider a  logic $\CSL$  obtained by adding to a  propositional language just one binary modal connective
$\csim$ expressing comparative similarity. In this language the above examples can be encoded
(using a description logic notation) by:
\beq $(1) Reddish \equiv \{Red\} \clsr \{Green, \ldots,  black\}$\\
(2) $Clio \sqset (Peugeot207 \clsr Golf)$
\enq
In a more general setting, the language might contain several $\clsr_{Feature}$ where each $Feature$ corresponds to a specific 
distance function $d_{Feature}$ measuring the similarity of objects with respect to one $Feature$ (size, price, power, taste, color...).
In our setting  a KB about cars may collect assertions of the form (2) and others,  say:
\beq 
(3) $Clio \sqset (Golf  \clsr Ferrari430)$\\
(4) $Clio \sqset (Peugeot207  \clsr MaseratiQP)$
\enq
together with some general axioms for classifying cars:
\beq 
$Peugeot207 \sqset Citycar$\\
$SportLuxuryCar \equiv MaseratiQP \mun Ferrari430$
\enq
Comparative similarity assertions such as (2)--(4) might not necessarily be  the fruit of an objective numerical calculation of similarity measures, but they could  be determined  just by  the (integration of) subjective opinions  of  agents,  answering, for instance, to questions like:   "Is Clio more similar to Golf or to Ferrari 430?"'. 
In any case, the logic $\CSL$ allows one to perform some kind of reasoning, for instance the following conclusions will be supported: 
\beq 
$Clio \sqset (Peugeot207  \clsr Ferrari430)$\\
$Clio \sqset (Citycar \clsr SportLuxuryCar)$
\enq
and also $Clio \sqset (Citycar \clsr SportLuxuryCar  \mint 4Wheels)$.

The semantics of $\CSL$ is defined in terms of distance spaces, that is  to say
structures equipped by a distance function $d$, whose properties may vary
according to the logic under consideration. In this setting, the evaluation of
$A \csim B$ can be informally stated as follows:
$x \in A \csim B$ iff $d(x, A) < d(x, B)$
meaning that the object $x$ is an instance of the concept $A \csim B$ (i.e. it
belongs to things that are more similar to $A$ than to $B$) if $x$  is 
strictly closer to $A$-objects than to $B$-objects according to distance
function $d$,  where the distance of an object to a set of objects is defined
as the \emph{infimum} of the distances to each object in the set.

In \cite{Wolter:LPAR05,Wolter08,conf/birthday/KuruczWZ05,DBLP:journals/logcom/SheremetTWZ07}, the authors have investigated the logic $\CSL$ with respect to
different classes of distance models, see \cite{Wolter08} for a survey of results about
decidability, complexity, expressivity, and axiomatisation. Remarkably it is
shown that $\CSL$ is undecidable over subspaces of the reals.  Moreover $\CSL$ over arbitrary distance spaces can be
seen as a fragment, indeed a powerful one (including for instance the logic
$\mathbf{S4}_u$ of topological spaces),  of a general  logic for spatial reasoning
comprising  different modal operators defined by  (bounded)
quantified distance expressions.

The authors have pointed out that in case the distance spaces are assumed to be
 \emph{minspaces}, that is spaces where the infimum of a set of distances is actually their
\emph{minimum}, the logic $\CSL$ is naturally related to some conditional logics.  The
semantics of  the latter is often expressed in terms of
preferential structures, that is to say  possible-world structures equipped by a family of
strict partial (pre)-orders $\prec_x$ indexed on objects/worlds \cite{Lewis:73,Stalnaker68}. The intended meaning of the relation  
$y \prec_x z$ is  namely that $x$ is more similar to $y$ than to $z$. 
It is not hard to see that the semantics over minspaces is equivalent to the semantics over preferential structures satisfying the well-known principle of Limit Assumption according to which the set of minimal elements of a non-empty set always exists.

The  minspace property entails the restriction  to  spaces where the  distance function is discrete.
This requirement does not seem incompatible with the purpose of representing qualitative similarity comparisons, whereas it might not be reasonable for applications of $\CSL$ to spatial reasoning.

In this paper we contribute to the study of $\CSL$ over minspaces. 
We first show (unsurprisingly) that the semantics of $\CSL$ on minspaces can be equivalently restated in terms of
preferential models satisfying some additional conditions, namely modularity,  centering, and limit assumption. 
We then give a \emph{direct} axiomatization of this logic. This problem was not considered in detail in \cite{Wolter08}. In that paper 
an axiomatization of CSL over arbitrary distance models is proposed, but it makes use of an  additional operator. Our  axiomatisation is simpler and only employs $\clsr$.
Next, we define  a tableaux calculus for checking satisfiability of $\CSL$ formulas. Our tableaux procedure makes use of labelled formulas and pseudo-modalities indexed on worlds $\Box_x$, similarly to the calculi for conditional logics defined in \cite{DBLP:conf/tableaux/GiordanoGOS03,GGOS11}. Termination is assured by suitable  blocking conditions.  To the best of our knowledge our calculus provides the first known practically-implementable decision procedure for $\CSL$ logic.

\section{The logic of \emph{Comparative Concept Similarity} $\CSL$}\label{section:csl}

The language $\Lcsl$ of $\CSL$ is generated from a set of propositional variables $V_i$ by ordinary propositional connectives plus $\clsr$: 
  $A,B ~::=~ V_i ~|~ \neg A ~|~ A \sqcap B ~|~ A \clsr B$.

The semantics of $\CSL$ introduced in~\cite{Wolter:LPAR05} makes use of \emph{distance spaces} in order to represent the similarity degree between objects. A distance space is a pair $(\Delta,d)$ where $\Delta$ is a non-empty set, and $d : \Delta \times \Delta \rightarrow \mathds{R}^{\geq 0}$ is a \emph{distance function} satisfying the following condition:
  \begin{equation}
    \tag{ID}\label{eq:ID} \forall x, y \in \Delta, ~~d(x,y) = 0 \text{ iff } x = y
  \end{equation}
Two further properties are usually considered: symmetry and triangle inequality. We briefly discuss them below.

The distance between  an object $w$ and a non-empty subset $X$ of $\Delta$ is defined by $d(w,X) = \inf\{d(w,x) ~|~ x \in X \}$. If $X=\emptyset$, then $d(w,X) = \infty$.
If for every object $w$ and for every (non-empty) subset $X$ we have the following property
\begin{equation}
  \tag{MIN} \label{eq:min} \inf\left\{d(w,x) ~\middle|~ x \in X \right\} = \min\left\{d(w,x) ~\middle|~ x \in X \right\},
\end{equation}
we will say that $(\Delta,d)$ is a \emph{minspace}.

We next define $\CSL$-distance models as  Kripke models based on distance spaces:
\begin{definition}[$\CSL$-distance model]\label{def:distmodel}
  A $\CSL$-distance model is a triple $\M = (\Delta,d,.^\M)$ where:
  \begin{itemize}
    \item $\Delta$ is a non-empty set of \emph{objects}.
    \item $d$ is a distance on $\Delta^\M$ (so that $(\Delta,d)$ is a distance space).
    \item $.^\M : \Vp \rightarrow 2^\Delta$ is the \emph{evaluation function} which assigns to each propositional variable $V_i$ a set $V_i^\M \subseteq \Delta$. We  further stipulate:      
    \begin{quote}
    $\bot^\M = \emptyset$\ \ \ \ \ \ \ \ $(\neg C)^\M = \Delta - C^\M$\ \ \ \ \ \ \ \ $(C \sqcap D)^\M = C^\M \cap D^\M$\\
    $(C \clsr D)^\M = \left\{w \in \Delta ~\middle| d(w,C^\M) < d(w,D^\M) \right\}.$
    \end{quote}
  \end{itemize}

  If $(\Delta, d)$ is a minspace, $\M$ is called a \emph{$\CSL$-distance minspace model} (or simply a \emph{minspace model}).
  We say that a formula $A$ is \emph{valid in a model} $\M$ if $A^\M = \Delta$. We say that a formula $A$ is \emph{valid} if $A$ is valid in every $\CSL$-distance model.
\end{definition}

As mentioned above, the distance function might be required to satisfy the further conditions of symmetry $(SYM)$ ($d(x,y) = d(y,x)$) and triangular inequality $(TR)$ ($d(x,z) \leq d(x,y) + d(y,z)$). It turns out that $\CSL$ cannot distinguish between minspace models which satisfy $(TR)$ from models which do not. In contrast \cite{Wolter:LPAR05}, $\CSL$ has enough expressive power in order to distinguish between symmetric and non-symmetric minspace models. 
As a first step, we concentrate here on the general non-symmetric case, leaving the interesting symmetric case to further research. 

$\CSL$ is a logic of pure qualitative comparisons. This motivates an alternative semantics where the distance function is replaced by a family of comparisons relations, one for each object. We call this semantics \emph{preferential} semantics, similarly to the semantics of conditional logics \cite{Nute:80,Lewis:73}.  Preferential structures are equipped by a family of strict pre-orders. We may interpret this relations as expressing a comparative similarity  between objects. For three objects, $x\prec_w y$ states that $w$ is more similar to $x$ than to $y$.

The preferential semantics in itself is more general than distance model semantics. However, if we assume the additional conditions of the definition~\ref{def:prefprop}, it turns out that these two  are equivalent (theorem \ref{th:semantic-equivalence}).

\begin{definition}\label{def:prefprop}
  We will say that a preferential relation $\prec_w$ over $\Delta$:
  \begin{enumerate}
  \renewcommand{\labelenumi}{(\roman{enumi})}
    \item is \emph{modular} iff $\forall x,y,z \in \Delta$, $(x \prec_w y) \impl (z \prec_w y ~\vee~ x \prec_w z)$.
    
    \item is \emph{centered} iff $\forall x \in \Delta$, $x = w ~\vee~ w \prec_w x$.

    \item satisfies the \emph{Limit Assumption} iff $\forall X \subseteq \Delta$, $X \neq \emptyset ~\impl~ \min_{\prec_w}(X) \neq \emptyset$.\footnote{We note that the Limit Assumption implies that the preferential relation is asymmetric. On the other hand, on a finite set, asymmetry implies Limit Assumption. Modularity and asymmetry imply that this relation is also transitive and irreflexive.}
    where $\min_{\prec_w}(X) = \{y\in X \mid \forall z\in \Delta (z \prec_w y \rightarrow z\notin X)\}$.
  \end{enumerate}
\end{definition}

Modularity is strongly related to the fact that the preferential relations represents distance comparisons. This is the key property to enforce the equivalence with distance models. Centering states that $w$ is the \emph{unique} minimal element for its preferential relation $\prec_w$, and can be seen as the preferential counterpart of~\eqref{eq:ID}. The Limit Assumption states that each non-empty set has at least one minimal element wrt. a preferential relation (i.e it does not contain an infinitely descending chain), and corresponds to \eqref{eq:min}.

\begin{definition}[$\CSL$-preferential model]\label{defi:prefsem}
  A $\CSL$-preferential model is a triple $\M = (\Delta,(\prec_w)_{w\in\Delta}, .^\M)$ where:
  \begin{itemize}
    \item $\Delta^\M$ is a non-empty set of \emph{objects} (or \emph{possible worlds}).

    \item $(\prec_w)_{w\in\Delta}$ is a family of \emph{preferential relation}, each one being modular, centered, and satisfying the \emph{limit assumption}.
    
    \item $.^\M$ is the evaluation function defined as in definition \ref{def:distmodel}, except for $\clsr$:
      \begin{equation*}
        (A \clsr B)^\M = \left\{ w \in \Delta \middle| \exists x \in A^\M \text{ such that } \forall y \in B^\M, ~x \prec_w y \right\}
      \end{equation*}
  \end{itemize}
  Validity is defined as in definition \ref{def:distmodel}.
\end{definition}

We now show the equivalence between preferential models and distance minspace models. We say that a $\CSL$-preferential model $\I$ and a $\CSL$-distance minspace model $\J$ are \emph{equivalent} iff they are based on the same set $\Delta$, and for all formulas $A \in \Lcsl$, $A^\I = A^\J$.

\begin{theorem}[Equivalence between $\CSL$-preferential models and $\CSL$-distance models]\label{th:semantic-equivalence}
  \begin{enumerate}
    \item For each $\CSL$-distance minspace model, there is an equivalent $\CSL$-preferential model.
    \item For each $\CSL$-preferential model, there is an equivalent $\CSL$-distance minspace model.
  \end{enumerate}
\end{theorem}

\begin{proof}
  \begin{enumerate}
    \item (\cite{Wolter:LPAR05}):  
    given $\I = (\Delta^\I,d,.^\I)$ a $\CSL$-distance minspace model, just define a preferential model $\J$ by stipulating $x\prec_w y$ iff $d(w,x)<d(w,y)$, and for all propositional variable $V_i$, $V_i^\J =V_i^\I$. It is to check  that $\prec_w$ is modular, centered, and satisfies the limit assumption, and that $\I$ and $\J$ are equivalent.
    
    \item Since the relation  $\prec_w$ is modular, we can assume that there exists a \emph{ranking function} $r_w: \Delta \to \mathds{R}^{\geq 0}$ such that $x \prec_w y$ iff $r_w(x) < r_w(y)$. Therefore, given a $\CSL$-preferential model $\J = (\Delta^\J,(\prec_w)_{w\in\Delta^\J},.^\J)$, we can define a $\CSL$-distance minspace model $\I = (\Delta^\J,d, .^\J)$, where the distance function $d$ is defined as follow: if $w=x$ then $d(w,x) = 0$, and $d(w,x) = r_w(x)$ otherwise.
    We can easily check that (i) $I$ is a minspace because of  the limit assumption, and that (ii) $\I$ and $\J$ are equivalent; this is proved by induction on the complexity of formulas.
  \end{enumerate}

\end{proof}

We have mentioned the relation with conditional logics. These logics, originally introduced by Lewis and Stalnaker \cite{Lewis:73,Stalnaker68}, contain a connective $A > B$ whose reading is approximatively "`if A were true then B would also be true"'\footnote{To this regard, in alternative to the concept/subset interpretation mentioned so far, the  formula $A \clsr B$ may perhaps be  read as  "`$A$ is (strictly) more plausible than $B$"'. This interpretation may intuitively explain the relation with the conditional operator.}. The idea is that a world/state $x$ verifies $A > B$ if $B$ holds in all states $y$ that are most similar to $x$ that is:
\beq 
$x\in A > B^\M$ iff $\min_{\prec_x}(A^\M) \subseteq B^\M$
\enq
The two connectives $\clsr$ are interdefinable as shown in \cite{Wolter:LPAR05}:
$$A > B \equiv  (A \clsr (A \land \lnot B)) \lor  \lnot (A \clsr \bot)$$
$$A \clsr B \equiv ((A \lor B) > A) \land (A > \lnot B) \land \lnot (A > \bot)$$
By means of this equivalence, an (indirect) axiomatization of $\clsr$ can be obtained: just take an  axiomatization of the suitable conditional logic (well known) and add the definition above. On the other hand an axiomatisation of $\CSL$ over arbitrary  distance models is presented in \cite{Wolter08}, however it makes use of an extended language, as we comment below.  Moreover, the case of minspaces has not been studied in details. 
Our axiomatisation is contained in  fig. \ref{fig:axiomes}. 
\begin{figure}[ht!]
  \begin{center}
  \scalebox{0.85}{
    \(
      \begin{array}{ll@{\hspace{3em}}ll}
              \nmr{(1)}{0}
      &
        \neg( A \clsr B) \sqcup \neg( B \clsr A)
      &
        \nmr{(2)}{0}
      &
        (A \clsr B) \impl (A \clsr C) \sqcup (C \clsr B)
      \\
      \\
        \nmr{(3)}{0}
      &
        A \sqcap \neg B \impl (A \clsr B)
      &
        \nmr{(4)}{0}
      &
        (A \clsr B) \impl \neg B 
      \\
      \\
        \nmr{(5)}{0}
      &
        (A \clsr B) \sqcap (A \clsr C) \impl (A \clsr (B\sqcup C))
      &
      \nmr{(6)}{0}
      &
         (A \clsr \bot)   \rightarrow \lnot (\lnot (A \clsr \bot)\clsr \bot)
      \\
      \\
        \nmr{(Mon)}{5}
      &
        \shortstack{$\vdash(A \impl B)$ \\ 
          \mbox{}\hrulefill\mbox{}\\
          $\vdash(A \clsr C) \impl (B \clsr C)$
        }
      &
        \nmr{(Taut)}{0}
      &
        \shortstack{Classical tautologies and rules.}
      \\
      \\
      \end{array}
    \)
    }
  \end{center}
  \caption{$\CSMS$ axioms.}\label{fig:axiomes}
\end{figure}
The axioms $(1)$ and $(2)$ capture respectively the asymmetry and modularity of the preference relations, whereas $(3)$ and $(4)$ encode centering and the  minspace property. By $(5)$, we obtain that $\clsr$ distributes over disjunction on the second argument, since the opposite direction is derivable. 
The axiom $(6)$ is similar to axiom $(33)$ of the axiomatization  in  \cite{Wolter08}, it says that the modality   $\Diamond A \equiv A \clsr \bot$ has the properties of S5. Finally, the rule $(Mon)$ states the monotonicity of $\clsr$ in the first argument, a dual rule stating the anti-monotonicity in the second argument is derivable as well. 

The axiomatisation of $\CSL$ provided in \cite{Wolter08} for arbitrary distance spaces makes use of the operator $\tobedef A$  that, referring to  preferential models, selects elements $x$ for which $\min_{\prec_x}(A)$ is non-empty. As observed in \cite{Wolter08},  an axiomatization of $\CSL$ over minspaces can then be obtained by just adding the axiom $\tobedef A \leftrightarrow (A \clsr \bot)$. However our axiomatization is significantly simpler (almost one half of the axioms). 

We can show that our axiomatization is sound and complete with respect to the preferential semantics, whence wrt minspace models (by theorem \ref{th:semantic-equivalence}).

\longreport{
\begin{theorem}[See \cite{TechReport}]
   A formula is derivable in $\CSMS$ iff it is valid in every $\CSL$-preferential model. 
\end{theorem}
}{

\begin{theorem}
   A formula is derivable in $\CSMS$ iff it is valid in every $\CSL$-preferential model. 
\end{theorem}

 \noindent The following theorems and inference rule are derivable from the axioms:

\begin{description}\label{theo}
\item[T1]  $A \rightarrow (A\clsr \bot)$ by (3)
\item[T2]  $\neg(A \clsr A)$ by (1)
\item[T3]  $\neg(A \clsr \top)$ by (2)
\item[T4]  $( (A\clsr \bot) \clsr \bot)  \rightarrow  (A\clsr \bot)$   by T1 and (6)
\item[T5]  $ (A\clsr B) \sqcap   (B\clsr C)  \rightarrow  (A\clsr C) $  by (1) and (4)
\item[T6]    $\forall n > 0, ~ \vdash (A \clsr B_1) \sqcap \ldots \sqcap (A \clsr B_n) \impl (A \clsr (B_1 \sqcup \ldots \sqcup B_n))$ by induction over $n$ and (5)
    \item[T7] $\forall n > 0, ~ \vdash (A \clsr B_1) \sqcap \ldots \sqcap (A \clsr B_n) \impl ((A \sqcap \neg B_1 \sqcap \ldots \sqcap \neg B_n) \clsr (B_1 \sqcup \ldots \sqcup B_n))$
  \item[R1]  If $\vdash(A \impl B)$ then   $\vdash(C \clsr B) \impl (C \clsr A)$ by (1) and RM

\end{description}

\noindent  Theorem  (T1) corresponds to the $T$-axiom $A \rightarrow \Diamond  A$; axiom (6) is the S5 axiom (Euclidean)  $\Diamond A \rightarrow \Box \Diamond A$. Hence making use of  (2) and (6) we can derive the S4 axiom  (T4).


\noindent  We can show that our axiomatisation is sound and complete with respect to the preferential semantics introduced above.

  \begin{theorem}[Soundness of $\CSMS$]\label{Soundn} If a formula is derivable in $\CSMS$,
then it is  $\CSMS$-valid.
\end{theorem}

\begin{theorem}[Completeness of $\CSMS$]\label{complete}
 If a formula is $\CSMS-$valid, then it is derivable in $\CSMS$.
\end{theorem}

\noindent  Soundness is straightforward. We show that every axiom is $\CSL$-valid.  

\noindent  The completeness is shown by the construction of a canonical
model.  We define consistent and maximal consistent formula sets in the usual way:

\begin{definition}

\begin{itemize}
 \item[]
 \item A set of formulas $\Gamma$ is called
\emph{inconsistent} with respect to $\CSMS$ iff there
is a finite subset of~~$\Gamma$, $\{A_1, \dots  A_n\}$
such that $\vdash_{\CSMS} \neg  A_1\sqcup \neg  A_2 \sqcup
\dots \neg  A_n$. $\Gamma$ is called \emph{consistent} if
$\Gamma$ is not inconsistent. If an (in)consistent ~~$\Gamma$
contains only one formula $ A$, we say that $ A$ is
(in)consistent.
\item A set of formulas $\Gamma$ is called \emph{maximal consistent}
iff it is consistent and if for any formula $ A$ not in
$\Gamma$, $\Gamma \cup \{ A\}$ is inconsistent. \end{itemize}
\end{definition}
We will use  properties of maximal consistent  sets, the proofs of
which can be found in most textbooks of logic. In particular:

\begin{lemma}\label{theo:max} Every consistent set of formulas is contained in a maximal consistent set of formulas.
\end{lemma}

\begin{lemma}\label{lem:max}
Let $w$ be a maximal consistent set of formulas and $ A$,  $
B$ formulas in $\Lcsl$. Then $w$ has the following properties:

\begin{enumerate}
\item If $\vdash_{\CSMS}  A \rightarrow  B$ and $ A \in w$, then $ B\in w$
\item If from $ A\in w$ we infer $ B\in w$, then $ A\rightarrow  B\in
w$.
\item  $ A\sqcap  B \in w$ iff $ A\in w$ and $ B\in w$
\item $ A  \not\in w$ iff $\neg  A \in w$
 \end{enumerate}
 \end{lemma}

\noindent Let $\U$ be the set of all maximal consistent sets.

 \begin{definition}
Let $x,y$ be maximal consistent formula sets and $A, B$ be $\Lcsl$-formulas. We define  \begin{description}
\item[1] $R(x,y) {\mbox{ iff }}  \forall A \in \Lcsl     {\mbox{ if }}   A \in y   {\mbox{ then  }}  (A \clsr \bot) \in x$
\item[2]  $w^A = \{ \neg B  \mid  (A \clsr B) \in w\}$
\end{description}
 \end{definition}
  
  \begin{property}
$R$ is an equivalence relation.
\begin{proof}
\begin{description}
\item[1]  $R$ is reflexive by $T1$
\item[2]  $R$ is transitive by $T4$
\item[3]  $R$ is symmetric by axiom $(6)$
\end{description}

\end{proof}
\end{property}

\noindent  For $x\in \U$, we note $ \tilde{x}$ the  equivalence class of $x$  with respect to $R$.
      
\noindent  The following properties hold for  $w^A$:

\begin{lemma}\label{le:propwA}
\begin{enumerate}
\item  If $\{A\}$  is consistent, then $w^A \cup \{A\}$  is consistent.
\item   $(A \clsr B)\in w$ iff $\forall x $ if $w^A \subseteq x$ then $\neg B \in x$.
\item $w^A \subseteq w $
\item   If $\{A\}$  is consistent and  $ (A \clsr \bot )\in w$ then $\exists z \in \Delta $ such that $A \in z$ and $w^A \subseteq z$
\end{enumerate}

\begin{proof}
\begin{enumerate}
\item 
Suppose that  $w^A \cup \{A\}$  is inconsistent. Then there are formulas  $\neg B_1, \dots \neg B_n \in w^A$ such that 
$\vdash_{\CSMS} B_1  \sqcup  \dots \sqcup B_n \sqcup \neg A$.
We can then derive
\begin{description}
\item[(i)] $\vdash_{\CSMS} A \rightarrow B_1  \sqcup  \dots \sqcup B_n $
\item[(ii)] $\vdash_{\CSMS} (A \clsr (B_1  \sqcup  \dots \sqcup B_n)) \rightarrow (A \clsr A) $ from (i) by R1
\item[(iii)] $\vdash_{\CSMS} \neg( (A \clsr   B_1)  \sqcap  \dots \sqcap (A\clsr B_n)) $ (from (ii) by T2 and axiom (5)
Contradiction with the consistency of $w$ since all $(A \clsr   B_i) \in w$. 
\end{description}

\item  ``$\Rightarrow$'' immediately by definition of $w^A$\\
``$\Leftarrow$'' We first show that $w^A \cup \{B\}$ is inconsistent. Suppose that this is not the case. Then there is $z\in \U$ and $w^A \cup \{B\} \subseteq z$. Hence $\neg B \in z$, from  the precondition. But $B\in z$, contradicting  the consistency of $z$. Since $w^A \cup \{B\}$ is inconsistent, there are formulas $\neg B_1, \dots \neg B_n \in w^A$ and   $\vdash_{\CSMS} \neg B_1  \sqcap  \dots \sqcap \neg B_n \rightarrow \neg B$. We can then derive

\begin{description}
\item[ (i)] $\vdash_{\CSMS} B  \rightarrow    B_1  \sqcup  \dots \sqcup   B_n $
\item[ (ii)] $\vdash_{\CSMS} (A\clsr      (B_1  \sqcup  \dots \sqcup   B_n)  \rightarrow  (A\clsr B) $ from (i) by R1
\item[ (iii)] $\vdash_{\CSMS} ((A\clsr   B_1)   \sqcap  (A\clsr   B_1) \dots \sqcap   (A\clsr B_n))  \rightarrow  (A\clsr B) $ from (ii) by (T6)

\end{description}
Since $(A\clsr   B_i) \in w$, we conclude $ (A\clsr B)\in w$
\item  immediate by axiom (2).
\item By 1, we have that $w^A \cup \{A\}$  is consistent, hence is is contained in a maximal consistent formula set $z\in \U$ by lemma \ref{theo:max}. We show then that $w^A$ is contained in a set $x\in \Delta$
We  show that for all $x\in \U$, if $w^A \subseteq x$ and  $(A\clsr \bot) \in w$, then    $R(w,x)$. 
We have    $\forall C \in w, (C\clsr \bot) \in w$, because of the reflexivity of $R$. 
By axiom (4) we have $(A\clsr\bot) \rightarrow (A\clsr \neg(C\clsr \bot))\sqcup (\neg(C\clsr \bot) \clsr \bot) $. By axiom  (6), we obtain $(\neg(C\clsr \bot) \clsr \bot) \not\in w$, since $(C\clsr \bot) \in w$. Hence $(A\clsr \neg(C\clsr \bot)) \in w$. This entails $(C\clsr \bot) \in w^A$ and therefore $(C\clsr \bot) \in x$.  This means that we have  $R(w,x)$.   
 Hence we have $x\in \Delta$ and $w^A \subseteq x$.


\end{enumerate}

\end{proof}
\end{lemma}

\noindent      We now are in a position to define the canonical model. 
 \begin{definition}[Canonical Model]\label{de:canonical}
 Since $C$ is not derivable in $\CSMS$, $\neg C$ is consistent, and so there is a maximal consistent set $z \in \U$ such that $\neg C\in z$. We define the   canonical model   $\Mc = (\Delta,(\prec_w)_{w\in\Delta},.^\Mc)$ as follows:
    \begin{itemize}
      \item $\Delta =  \tilde{z}$.
      \item $x\prec_w y$ iff there exists a formula $B \in y$ such that for all formulas $A \in x$, $(A \clsr B) \in w$.
      \item $V_i^\Mc = \{ x \in \Delta ~|~ V_i \in x \}$, for all propositional variables $V_i$.
    \end{itemize}

\end{definition}

For $A\in \Lcsl$, we define the set of objects in $\Delta$  containing $A$ by $ \|A\| = \{ z \mid A \in z\cap \Delta\}$.

   \begin{lemma}
For each  $w\in\Delta$,   $\prec_w$ is   centered and modular.
\begin{proof}
\begin{description}
 
\item[1]  $\prec_w$ is  centered. Let be $x\neq y$. Then there is (i) $B\in y$ and $B \not\in x$, i.e. $\neg B \in x$. Let be any $\Lcsl$-formula $A$ with $A\in x$. Then $A\sqcap \neg B \in x$ from  which follows that  (ii)$(A\clsr B) \in x$ by axiom (3). From (i) and (ii) we obtain $x\prec_x y$.
\item[2]  $\prec_w$ is modular.  Let be $x\prec_w y$ and suppose there is $u\in \Delta $ such that  $x \not  \prec_w u$ and $u \not \prec_w y$. We get then: 

\begin{description}
\item[(i)]  $\exists B \in y$ such that   $\forall A \in x, ~  \csfw{A}{B}{w}$
\item[(ii)]    $\forall C\in u~  \exists A' \in x$  such that  $\neg(A'\clsr C) \in w$  and 
\item[(iii)]    $\forall B'\in y ~  \exists C' \in u$ such that  $\neg(C'\clsr B') \in w$
\end{description}
Then we have  $\csfw{A'}{B}{w}$ from (1),  $\neg(A'\clsr C') \in w$  from (ii) and    $\neg(C'\clsr B) \in w$ from (iii). By axiom (1) and (ii), we get   $\csfw{C}{A'}{w}$ and by transitivity (T5) and (i), we obtain  $\csfw{C'}{B}{w}$ which contradicts the consistency of $w$. 

 \end{description}

\end{proof}

\end{lemma}

\noindent  Subsequently, we show a weak variant of  the limit assumption for sets of objects  satisfying a formula.

\begin{lemma}\label{lem:limitassumption}
If $\|A\|  \neq \emptyset $ then $min_{\prec_w}(\|A\|) \neq \emptyset$.

\begin{proof}


Let be $\|A\|  \neq \emptyset $. Then $w^A \cup \{A\}$  is consistent by lemma  \ref{le:propwA}.1. Therefore there is a maximal consistent set $z\in \U$ such that  $w^A \cup \{A\} \subseteq z$.  We show that $z \in min_{\prec_w}( \|A\|)$.     Suppose for the contrary that  there is $y \in A^{Mc}$ and $y\prec_w z$. By definition of canonical model, this means that $\exists  C \in z$ such that  $\forall B \in y$,  $  (B\clsr C) \in w$. Since $A \in y$, we have  
$(A \clsr C) \in w$, hence $\neg C \in w^A$, which entails $\neg C \in  z$ contradicting the consistency of z. 
\end{proof}

It is not hard to see  that a formula $A$ is valid wrt. the weak variant of the limit assumption iff  it is valid wrt. the strong variant.

\end{lemma}

\begin{lemma}\label{le:complmod}
$ \|A\| = A^{Mc}$
\begin{proof}

The proof is by induction on the construction of formulas.

\begin{description}
\item For atomic $V_i$ it follows from the model definition. For classical formulas the proof is standard.

\item "$\Rightarrow$": Let be $F \in  w$, $F = (A\clsr B)$.  
 $w^A$ is consistent by lemma  \ref{le:propwA}, 3.  Let be $x\in \Delta$, such that $w^A \in x$ (\ref{le:propwA}, 4).  By axiom (4), we have  $((A\clsr C)\sqcup (C\clsr B) \in w$ for any  $C\in x$. By lemma \ref{le:propwA}, 2, we have then    $(A\clsr C) \not\in w$, hence  $\neg(A\clsr C) \in w$ from which we get $(C\clsr B) \in w$    $\forall C \in x$. By the definition of $\prec_w$ we have then  if for all $y\in \Delta$, if  $B \in y$ then  $x\prec_w y$. This means that $w\in (A\clsr B)^\Mc$

\item"$\Leftarrow$":  Let be $w\in  (A\clsr B)^\Mc$. Then   $A^\Mc \neq \emptyset$. By \IV, $A\in x$ for all $x \in A^\Mc$. Then $A$ is consistent and since $(w,x)\in R$, $(A\clsr \bot)\in w$. By  lemma \ref{lem:max}.1, and axiom (4), we have  $$(i)(A\clsr B) \sqcup (B\clsr \bot) \in w  $$ By lemma \ref{le:propwA}.4, there is $z\in \Delta$ such that $w^A \subseteq z$. We consider two cases:
\begin{description}
\item[(a)] $B^\Mc = \emptyset$. If $B$ is inconsistent, we have trivially $(A\clsr B ) \in w$,  by  $(A\clsr \bot)\in w$ and (RM), R1. If $B$ is not inconsistent, we observe that $$(ii) \neg(B\clsr \bot) \in w $$ if not,  by lemma \ref{le:propwA}.4, there would be $z' \in \Delta$ with $B\in z'$, i.e. $z' \in B$ by \IV which is impossible  since $B^\Mc = \emptyset$. We conclude  $(A\clsr B) \in w$ by (i). 
\item[(b)] $B^\Mc \not= \emptyset$, since $B$ is consistent, there is $z'\in \Delta$  such that $w^B \subseteq z'$ and $B \in z'$. By  $w\in  (A\clsr B)^\Mc$  we have $z \prec_w z'$ ($z $ is minimal for $\prec_w$ in $\|A\|$). This means that  there is $C\in z'$ such that for all $D\in z, ~ (D\clsr C) \in w$. Since $A\in z$, we have $(A\clsr C)\in w$ from which we obtain $(A\clsr B)\sqcup (B\clsr C) \in w$. But we have $\neg(B\clsr C)\in w$ by lemma \ref{le:propwA}.2, from which we obtain $(A\clsr B)\in w$.
 
\end{description}

\end{description}

\end{proof}
\end{lemma}

By virtue of theorem \ref{th:semantic-equivalence}, we obtain:
\begin{corollary}\label{corol:semantic}
  $\CSMS$ is sound and complete wrt. the $\CSMS$-distance min-models.
\end{corollary}

}

\section{A Tableaux Calculus}\label{section:tableau}
In this section, we present a tableau calculus for $\CSL$, this calculus provides a decision procedure for this logic. 
We identify a tableau with  a set of sets of formulas $\Gamma_1,\ldots,\Gamma_n$. Each $\Gamma_i$ is called a \emph{tableau set}\footnote{A tableau set corresponds to a \emph{branch} in a tableau-as-tree representation.}.
Our calculus will make use of labels to represent objects of the domain. Let us consider  formulas $(A \clsr B)$ and $\neg (A \clsr B)$ under preferential semantics. We have:
  $$w \in (A\clsr B)^\M \text{ iff } \exists x (x \in A^\M \wedge \forall z (z \in B^\M \impl x \prec_w z))$$
In minspace models, the right part is equivalent to: 
  $$w \in (A \clsr B)^\M \mbox{ iff } \exists u \in A^\M \mbox{ and } \forall y(y \in B^\M \impl \exists x(x\in A^\M \wedge x \prec_w y) )$$

We now introduce a pseudo-modality $\square_w$ indexed on objects:
  $$x \in (\square_w A)^\M \text{ iff } \forall y(y \prec_w x \impl y \in A^\M)$$
 Its meaning is that $x \in (\square_w A)^\M$ iff $A$ holds in all  worlds preferred to $x$ with respect to $\prec_w$.  Observe  that  we have then  the  equivalence:
\begin{claim2}\label{claim:clsrsemantics}
   $w \in (A \clsr B)^\M \text{ iff } A^\M \neq \emptyset \mbox{ and } \forall y(y \notin B^\M \mbox{ or } y \in (\neg \square_w \neg A)^\M)$.
\end{claim2}

  This  equivalence will be used to decompose  $\clsr$-formulas in an analytic way.  The tableau rules make also use of a universal modality $\Box$ (and its negation).  The language of tableaux  comprises the following kind of formulas: $x:A, x:(\neg)\Box \neg A, x:(\neg)\Box_y \neg A, x <_y z$, where $x,y,z$ are labels and $A$ is a $\CSL$-formula. The meaning of $x: A$ is the obvious one: $x\in A^\M$. The reading of the rules is the following: we apply a rule
\begin{equation*}
  \shortstack{$\Gamma[E_1,\ldots,E_k]$ \\ 
    \mbox{}\hrulefill\mbox{}\\
    $\Gamma_1 ~|~ \ldots ~|~ \Gamma_n$}
\end{equation*}
to a tableau set $\Gamma$ if each formula $E_k$ is in $\Gamma$. We then replace $\Gamma$  with any tableau set $\Gamma_1,\ldots,\Gamma_n$.
As usual, we let $\Gamma, A$ stand for for $\Gamma \cup \{ A \}$, where $A$ is a tableau formula. The tableaux rules are shown in figure Figure \ref{fig:tabrule}. 
\begin{figure}[ht!]
    \begin{center}
    \scalebox{0.75}{
    \(
      \begin{array}{ll@{\hspace{2em}}ll}
      \\
        \nmr{(T\sqcap)}{5}
      &
         \shortstack{$\Gamma[x: A \sqcap B]$ \\ 
          \mbox{}\hrulefill\mbox{}\\
          $\Gamma,~ x: A,~ x: B$
        }
      &
        \nmr{(F\sqcap)}{6}
      &
        \shortstack{$\Gamma[x: \neg(A \sqcap B)]$ \\ 
          \mbox{}\hrulefill\mbox{}\\
          $\Gamma, ~x: \neg A ~|~\Gamma,~ x: \neg B$
        }
      \\
      \\
        \nmr{(NEG)}{5}
      &
        \shortstack{$\Gamma[x: \neg\neg A]$ \\ 
          \mbox{}\hrulefill\mbox{}\\
          $\Gamma,x: A$
        }
      &
        \nmr{(F1\clsr)}{6}
      &
        \shortstack{$\Gamma[x: \neg( A \clsr B)]$ \\ 
          \mbox{}\hrulefill\mbox{}\\
          $\Gamma, x:\square \neg A ~|~\Gamma, x:B ~|~ \Gamma, x:\neg A, ~x:\neg B$
        }
      \\
        \nmr{(T\clsr)(*)}{6}
      &
        \shortstack{$\Gamma[x: A \clsr B]$ \\ 
          \mbox{}\hrulefill\mbox{}\\
          $\Gamma, x:\neg \square\neg A,~ y: \neg B~|~\Gamma, y:B, y: \neg \square_x \neg A$
        }
      &
        \nmr{(F2\clsr)(**)}{6}
      &
        \shortstack{$\Gamma[x: \neg( A \clsr B), ~x:\neg A, ~x:\neg B]$ \\ 
          \mbox{}\hrulefill\mbox{}\\
          $ \Gamma, ~y:B,~ y:\square_x \neg A$
        }
      \\
      \\
      &
      &
        \nmr{(F1\square_x)}{6}
      &
        \shortstack{$\Gamma[z: \neg \square_x \neg A]$ \\ 
          \mbox{}\hrulefill\mbox{}\\
          $ \Gamma, x:\neg A ~|~ \Gamma, x: A$
        }
      \\
      \\        \nmr{(T\square_x)(*)}{5}
      &
        \shortstack{$\Gamma[z: \square_x \neg A, y <_x z]$ \\ 
          \mbox{}\hrulefill\mbox{}\\
          $\Gamma, ~y:\neg A, ~y: \square_x \neg A$
        }
      &
        \nmr{(F2\square_x)(**)}{6}
      &
        \shortstack{$\Gamma[z: \neg \square_x \neg A, ~x:\neg A]$ \\ 
          \mbox{}\hrulefill\mbox{}\\
          $\Gamma, ~y <_x z,~ y: A, ~ y:\square_x \neg A$
        }
      \\
      \\
        \nmr{(T\square)(*)}{5}
      &
        \shortstack{$\Gamma[x:\square \neg A]$ \\ 
          \mbox{}\hrulefill\mbox{}\\
          $\Gamma, ~y: \neg A, y:\square \neg A$
        }
      &
        \nmr{(F\square)(**)}{5}
      &
        \shortstack{$\Gamma[x:\neg\square\neg A]$ \\ 
          \mbox{}\hrulefill\mbox{}\\
          $\Gamma, ~y: A$
        }
      \\
      \\
        \nmr{(Mod)(*)}{6}
      &
        \shortstack{$\Gamma[z <_x u]$ \\ 
          \mbox{}\hrulefill\mbox{}\\
          $\Gamma, z <_x y ~|~ \Gamma, y <_x u$
        }
      &
        \nmr{(Cent)(***)}{5}
      &
        \shortstack{$\Gamma$ \\ 
          \mbox{}\hrulefill\mbox{}\\
          $\Gamma, x <_x y ~|~ \Gamma[x/y]$
        }
      \\
      \\
      \end{array}
    \)
  }  
  \end{center}
  (*) $y$ is a label occurring in $\Gamma$.
  (**)  $y$ is a new label not occurring in $\Gamma$.
  (***) $x$ and $y$ are two distinct labels occurring in $\Gamma$.
  
  \caption{Tableau rules for $\CSL$.}\label{fig:tabrule}
\end{figure}

Let us comment on the rules which are not immediately obvious. The rule for (T$\clsr$) encodes directly the semantics by virtue of claim \ref{claim:clsrsemantics}. However in the negative case the rule is split in two: if  $x$ satisfies $\lnot (A \clsr B)$, either $A$ is empty, or there must be an $y\in B$ such that there is no $z \prec_x y$ satisfying $A$;   if $x$ satisfies $B$ then $x$ itself fulfills this condition, i.e. we could take $y = x$, since $x$ is $\prec_x$-minimal (by centering). On the other hand, if $x$ does not satisfies $B$,  then $x$ cannot satisfy  $A$ either (otherwise $x$ would satisfy $A \clsr B$)   and there must be an $y$ as described above. This case analysis with respect to $x$ is performed by the (F1$\clsr$) rule, whereas the creation $y$ for the latter case is performed by (F2$\clsr$). We have a similar situation for the (F$\Box_x$) rule: let $z$ satisfy $\lnot \Box_x \lnot A$, then there must be an $y  \prec_x z$ satisfying $A$; but if $x$  satisfies $A$ we can take $x=y$, since $x\prec_x z$ (by centering). If $x$ does not satisfy $A$ then we must create a suitable $y$ and this is the task of the (F2$\Box_x$) rule. Observe that the rule does not simply create a $y\prec_x z$ satisfying $A$ but it creates a \emph{minimal} one. The rule is similar to the (F$\Box$) rule in modal logic GL (G\"{o}del-L\"{o}b modal logic of arithmetic provability) \cite{Boolos:93} and it is enforced by the Limit Assumption.  This formulation of the rules for (F$\clsr$) and for (F$\Box_x$) prevents the unnecessary creation of new objects whenever the existence of the objects required by the rules is assured by centering.
The rule $(Cent)$ is of a special kind: it has no premises (ie. it can always be applied) and generates two tableau sets: one with $\Gamma \cup \{x <_x y\}$, where $x$ and $y$ are two distinct labels occurring  in $\Gamma$), and one where we replace $x$ by $y$ in $\Gamma$, i.e. where we identify the two labels.

\begin{definition}[Closed set, closed tableau]\label{def:closed-branch}
  A tableau set $\Gamma$ is \emph{closed} if one of the three following conditions hold:
    (i) $x:A \in \Gamma$ and $x:\neg A \in \Gamma$, for any formula $A$, or $x:\bot \in \Gamma$.
    (ii) $y <_x z$ and $z <_x y$ are in $\Gamma$.
    (iii) $x:\neg \square_x A \in \Gamma$.

  A $\CSL$-tableau is closed if every tableau set is closed.
\end{definition}

In order to prove soundness and completeness of the tableaux rules, we introduce the notion of satisfiability of a tableau set  by a model.  

Given a tableau set $\Gamma$, we denote by $\mathrm{Lab}_\Gamma$ the set of labels occurring in $\Gamma$. 
\begin{figure}[ht!]
    \begin{center}
    \scalebox{0.60}{
      \psTree[treesep=100pt,levelsep=4cm]{\TR{$\{x:A, x:\neg B, x:\neg(A \clsr A)\}$}}
        \pstree{\TR{\shortstack{$\{x:A, x:\neg B, x:\neg(A \clsr A)$\\$x:\square\neg A\}$}}\tlput{$(F1\clsr)$}}
               {\TR{\shortstack{$\{x:A, x:\neg B, x:\neg(A \clsr A)$\\$x:\square\neg A$\\$x:\neg A\}$\\closed by def 6-(i)}}\tlput{$(T\square)$}}
        \pstree{\TR{\shortstack{$\{x:A, x:\neg B, x:\neg(A \clsr A)$\\$x:B\}$\\closed by def 6-(i)}}\tlput{$(F1\clsr)$}}{}
        \pstree{\TR{\shortstack{$\{x:A, x:\neg B, x:\neg(A \clsr A)$\\$x:\neg A\}$\\closed by def 6-(i)}}\trput{$(F1\clsr)$}}{}  
      \endpsTree
      }
  \end{center}
\caption{An example of tableau: provability of $A \sqcap \neg B \impl (A \clsr B)$.}
\end{figure}
\begin{definition}[$\CSL$-mapping, satisfiable tableau set]
  Let $\M = (\Delta^\M, (\prec_w)_{w\in\Delta^\M}, .^\M)$ be a preferential model, and $\Gamma$ a tableau set.
  A $\CSL$-mapping from $\Gamma$ to $\M$ is a function $f: \mathrm{Lab}_\Gamma \longrightarrow \Delta^\M$ satisfying the following condition:
  $\mbox{for every } y <_x z \in \Gamma, \mbox{ we have } f(y) \prec_{f(x)} f(z) \mbox{ in }\M$.
  
  Given a tableau set $\Gamma$, a $\CSL$-preferential model $\M$, and a $\CSL$-mapping $f$ from $\Gamma$ to $\M$, we say that $\Gamma$ is satisfiable under $f$ in $\M$ if
    $x:A \in \Gamma \mbox{ implies } f(x) \in A^\M.$
  A tableau set $\Gamma$ is satisfiable if it is satisfiable in some $\CSL$-preferential model $\M$ under some $\CSL$-mapping $f$.
  A $\CSL$-tableau is satisfiable if at least one of its sets is satisfiable.
\end{definition}

We can show that our tableau calculus is sound and complete with respect to the preferential semantics, whence with respect to minspace models (by theorem \ref{th:semantic-equivalence}).

\begin{theorem}[Soundness of the calculus]\label{th:tabsoundcomplete}
  A formula $A \in \Lcsl$ is satisfiable wrt. preferential semantics then any tableau started by $x:A$ is open.
\end{theorem}

\longreport{The proof of the soundness is standard, and can be found in \cite{TechReport}. In order to show completeness, we need the following definition:
}{
The proof of the soundness is standard: we show that rule application preserves satisfiability.

\begin{proof}[Soundness of the Tableau System]
  Let $\Gamma$ be a tableau set satisfiable in a $\CSL$-model $\M = \langle \Delta, (\prec_w)_{w\in\Delta}, .^\M \rangle$ under a $\CSL$-mapping $f$. We prove that if we apply one of the tableau rules to $\Gamma$, at least one of the new tableau sets generated by it is satisfiable. Moreover, a satisfiable set cannot be closed.
  
  \begin{itemize}
    \item The cases of $(T\sqcap)$, $(F\sqcap)$, $(T\square_x)$, $(T\square)$ and $(F\square)$ are easy and left to the reader. Soundness of $(F1\square_x)$ is trivial (it's a cut-like rule), as $(Mod)$ and $(Cent)$ which came from the modularity and centering property of the preferential relation $\prec_w$.
    
    \item $(T\clsr)$. Let $x:(A\clsr B) \in \Gamma$. For any label $y \in \mathrm{Lab}_\Gamma$, the application of this rule to $\Gamma$ will generate two tableau sets:
      $$\Gamma_1 = \Gamma \cup \{ y: \neg\square\neg A, y:\neg B \}$$
      $$\Gamma_2 = \Gamma \cup \{ y: B, y:\neg\square_x\neg A \}.$$
     where $y\in \mathrm{Lab}_\Gamma$.
     As $\Gamma$ is satisfiable in $\M$ under $f$, we have that $f(x) \in (A \clsr B)^\M$. By claim \ref{claim:clsrsemantics}, for all $y \in \mathrm{Lab}_\Gamma$ we have $f(y) \in (\neg\square\neg A)^\M$ (as $A^\M$ cannot be empty) and $f(y) \in (B \impl \neg \square_x \neg A)^\M$. We then have two cases:
     \begin{itemize}
     \item either $f(y) \in (\neg B)^\M$, and then $\Gamma_1$ is satisfiable.
     \item or $f(y) \in B^\M$ and then $f(y) \in (\neg \square_x \neg A)^\M$. $\Gamma_2$ is then satisfiable.
     \end{itemize}
     
    \item $(F1\clsr)$. Let $x:\neg(A \clsr B) \in \Gamma$.
      The rule will generate three tableau sets:
      $$\Gamma_1 = \Gamma \cup \{ x:\square\neg A\}$$
      $$\Gamma_2 = \Gamma \cup \{ x: B \} $$
      $$\Gamma_3 = \Gamma \cup \{ x:\neg A, x:\neg B \}$$ 
    
    As $\Gamma$ is satisfiable in $\M$ under $f$, we have $f(x) \in \neg (A \clsr B)^\M$. By claim \ref{claim:clsrsemantics}, we have that $f(x) \in \square (\neg A)^\M \mbox{ or } \exists y( y \in B^\M \mbox{ and } y \in (\square_x \neg A)^\M) $.
    
    If $f(x) \in (\square \neg A)^\M$, then $\Gamma_1$ is satisfiable. If not, we have two cases:
    \begin{itemize}
      \item either $f(x) \in B^\M$, and thus $\Gamma_2$ is satisfiable.
      \item either $f(x) \in (\neg B)^\M$, and then there is some object $y' \in \Delta$ such that $y'\in B^\M$ and $y' \in (\square_x \neg A)^\M$, so that $y' \neq f(x)$. Since the relation $\prec_x$ satisfies the centering property, we have $x \prec_x y'$. And from $y' \in (\square_x \neg A)^\M$, we can deduce that $f(x) \in (\neg A)^\M$, thus making $\Gamma_3$ satisfiable.
    \end{itemize}
    
    \item $(F2\clsr)$. This rule will generate the following set:
      $$\Gamma_1 = \Gamma \cup \{ y:\square_x \neg A, y:B \}$$
    where $y \notin \mathrm{Lab}_\Gamma$.
    As shown in the proof for $(F1\clsr)$, since $x:\neg( A\clsr B), x:\neg A, x:\neg B$ belong to $\Gamma$ and $\Gamma$ satisfiable in $\M$ by $f$, we have that there exists some $y' \in \Delta$ such that $y' \in (B \sqcap \square_x \neg A)^\M$. We construct a $CSL$-mapping $f'$ by taking $\forall u \neq y$, $f'(u) = f(u)$; and $f'(y) = y'$. It's then easy to show that $f'$ is a $\CSL$-mapping, and that $\Gamma_1$ is satisfiable in $\M$ under $f'$.
    
    \item $(F2\square_x)$. Let $z:\neg\square_x\neg A, x:\neg A$. This rule will generate the following tableau set:
      $$\Gamma_1 = \Gamma \cup \{ y <_x z, y:A, y:\square_x \neg A \}.$$
    where $y \notin \mathrm{Lab}_\Gamma$.
    Since $\Gamma$ is satisfiable in $\M$ under $f$, we have that $f(z) \in (\neg \square_x \neg A)^\M$. Therefore we have that $\exists y' \in \Delta$ such that $y' \in A^\M$ and $y' \prec_x f(z)$. We then let
    \begin{equation*}
      y'' \in \min_{\prec_x} \{ y' ~|~ y' \in \Delta ~\wedge~ y' \in A^\M ~\wedge~ y' \prec_x f(z) \}.
    \end{equation*}
    As $\prec_x$ satisfies the limit assumption, $y''$ exists, and it's easy to see that $y'' \in (\square_x \neg A)^\M$ (if not it would not be minimal).
    As $y \notin \mathrm{Lab}_\Gamma$, we define a new $\CSL$-mapping $f'$ by taking $\forall u \neq y, f'(u) = f(u)$, and $f(y) = y''$. It's then easy to check that $f'$ is indeed a $\CSL$-mapping, and that $\Gamma_1$ is satisfiable in $\M$ under $f'$.
  \end{itemize}

  Finally, we show that if $\Gamma$ is satisfiable, then $\Gamma$ is open. Suppose it is not. Then if $\Gamma$ is closed by def. \ref{def:closed-branch}-(i) or (ii), we immediately find a contradiction.
  Suppose that $\Gamma$ is closed by condition (iii), then  $x:\neg\square_x\neg A$ in $\Gamma$ for some formula $x:\neg\square_x\neg A$. Since $\Gamma$ is satisfiable, we would have that there is  $y \in \Delta$ such that $y \prec_{f(x)} f(x)$, obtaining a contradiction (by centering and asymmetry). 
\end{proof}

In order to show completeness, we need the following definition:
} 

\begin{definition}[Saturated tableau set]\label{def:saturation}
  We say that a tableau set $\Gamma$ is saturated if:
  \begin{description}
    \item[$(T\sqcap)$] If $x: A\sqcap B\in \Gamma$ then $x:A\in\Gamma$ and $x:B\in\Gamma$.
    
    \item[$(F\sqcap)$] If $x: \neg(A \sqcup B)\in\Gamma$ then $x:\neg A\in\Gamma$ or $x:\neg B\in\Gamma$.
    
    \item[$(NEG)$] If $x: \neg\neg A \in \Gamma$ then $x:A\in\Gamma$.
    
    \item[$(T\clsr)$] If $x:(A\clsr B) \in \Gamma$ then for all $y \in \mathrm{Lab}_\Gamma$, either $y:\neg B \in \Gamma$ and $\neg\square\neg A \in \Gamma$, or $y:B$ and $y:\neg\square_x\neg A$ are in $\Gamma$. 
    
    \item[$(F\clsr)$] If $x:\neg(A\clsr B) \in \Gamma$ then either (i) $x:\square\neg A \in \Gamma$, or (ii) $x:B \in \Gamma$, or (iii) $x:\neg A$ and $x: \neg B$ are in $\Gamma$ and there exists $y \in \mathrm{Lab}_\Gamma$ such that $y:B$ and $y:\square_x \neg A$ are in $\Gamma$.
    
    \item[$(T\square_x)$] If $z: \square_x \neg A\in \Gamma$ and $y <_x z \in \Gamma$, then $y:\neg A$ and $y:\square_x \neg A$ are in $\Gamma$.
    
    \item[$(F\square_x)$] If $z:\neg\square_x\neg A \in \Gamma$, then either (i) $x:A \in \Gamma$, or (ii) $x:\neg A \in \Gamma$ and there exists $y \in \mathrm{Lab}_\Gamma$ such that $y <_x z$, $y: A$ and $y: \square_x \neg A$ are in $\Gamma$.
    
    \item[$(T\square)$] If $x:\square \neg A \in \Gamma$, then for all $y \in \mathrm{Lab}_\Gamma$, $y:\neg A$ and $y: \square \neg A$ are in $\Gamma$.
    
    \item[$(F\square)$] If $x:\neg\square\neg A \in \Gamma$, then there is $y \in \mathrm{Lab}_\Gamma$ such as $y: A \in \Gamma$.

    \item[$(Cent)$] For all $x,y \in \mathrm{Lab}_\Gamma$ such that $x \neq y$, $x <_x y$ is $\in \Gamma$.
    
    \item[$(Mod)$] If $y <_x z \in \Gamma$, then for all labels $u \in W _\Gamma$, either $u <_x z \in \Gamma$, or $y <_x u \in \Gamma$.
  \end{description}
  We say that $\Gamma$ is saturated wrt. a rule $R$ if $\Gamma$ satisfies the corresponding saturation condition for $R$ of above definition.
\end{definition}

The following lemma shows that the preference relations $<_x$ satisfies the Limit Assumption for an open tableau set.

\longreport{
\begin{lemma}[See \cite{TechReport}]\label{lem:lim-assumption}
Let $\Gamma$ be an open tableau set containing only a finite number of
positive $\clsr$-formulas $x:  A_0 \clsr \ B_0$, $x:
 A_1 \clsr \ B_1$, $x:  A_2 \clsr \ B_2$, \dots,
$x:  A_{n-1} \clsr \ B_{n-1}$. Then $\Gamma$ does not contain
any infinite descending chain of labels $y_1<_x y_0$, $y_2<_x y_1$,
\dots, $y_{i+1}<_x y_i, \dots$.
\end{lemma}
}{\begin{lemma}\label{lem:lim-assumption}
Let $\Gamma$ be an open tableau set containing only a finite number of
positive $\clsr$-formulas $x:  A_0 \clsr \ B_0$, $x:
 A_1 \clsr \ B_1$, $x:  A_2 \clsr \ B_2$, \dots,
$x:  A_{n-1} \clsr \ B_{n-1}$. Then $\Gamma$ does not contain
any infinite descending chain of labels $y_1<_x y_0$, $y_2<_x y_1$,
\dots, $y_{i+1}<_x y_i, \dots$.
\end{lemma}
\begin{proof}
By absurdity, let   $\Gamma$ contain a descending chain of labels $\ldots,y_{i+1} <_x y_i <_x \ldots <_x y_1 <_x y_0$. This chain may only be generated by successive
applications of $(T\clsr)$ and $(F2\Box_x)$ to formulas $x:  A_i
\clsr \ B_i$ for $0 \leq i < n$. $\Gamma$ then contains the
following formulas for $0\leq i <n$: $ y_i: \neg \Box_x \neg
 A_i$,   $y_{i+1} <_x y_i$, $y_{i+1}:  A_i $,  $y_{i+1}: \Box_x
\neg  A_i$.

Here $(T\clsr)$ has been applied to every formula $x:  A_i
\clsr \ B_i$ once and with parameter $y_i$ previously (and
newly) generated by $(F2\Box_x)$ from $y_{i-1}: \neg \Box_x \neg
 A_{i-1}$.  The only way to make  the chain longer is by applying
$(T\clsr)$ a second time to one of the positive $\clsr$-formulas
labelled $x$ on $\Gamma$. Let this formula be $x:  A_k \clsr
\ B_k$ where $0 \leq k < n$.  Then $\Gamma$ contains further  $y_{n+1}:
 A_k$ (together with $y_n: \neg \Box_x \neg  A_k$, $y_{n+1} <_x
y_n$,  $y_{n+1}: \Box_x \neg  A_k $).

By the modularity rule, we get $~~y_{n+1} <_x y_{k+1}.$ Moreover,
$\Gamma$ contains also $y_{k+1}: \Box_x \neg  A_k,$ from which we
obtain by $(T\Box_x)$ $y_{n+1} : \neg  A_k$ so that $\Gamma$ is closed.
\end{proof}}


\begin{theorem}\label{th:complete1}
  If $\Gamma$ is an open and saturated tableau set, then $\Gamma$ is satisfiable.
\end{theorem}

\noindent \textbf{(Proof)}: Given an open tableau set $\Gamma$,  we define a canonical model $\M_\Gamma= \langle \Delta, (\prec_w)_{w\in\Delta}, .^{\M_\Gamma} \rangle$ as follows:
\begin{itemize}
  \item $\Delta = \mathrm{Lab}_\Gamma$ and $y \prec_x z$ iff $y <_y z \in \Gamma$.
  \item For all propositional variables $V_i \in \Vp$, $V_i^{\M_\Gamma} = \{ x ~|~ x:V_i \in \Gamma \}$
\end{itemize}

\noindent $\M_\Gamma$ is indeed a $\CSL$-model, as each preferential relation is centered, modular, and satisfies the limit assumption. The first two came from the rules $(Cent)$ and $(Mod)$, and we have the latter by lemma \ref{lem:lim-assumption}.

We now show that $\Gamma$ is satisfiable in ${\M_\Gamma}$ under the trivial identity mapping, i.e for all formula $C \in \Lcsl$:
 (i) if $x:C \in \Gamma$, then $x \in C^{\M_\Gamma}$.
 (ii) if $x:\neg C \in \Gamma$, then $x \in (\neg C)^{\M_\Gamma}$.

\begin{proof}
   We reason by induction on the complexity $cp(C)$ of a formula $C$, where we suppose that $cp((\neg)\square\neg A),cp((\neg)\square_x\neg A) < cp(A \clsr B)$.
    
  \begin{itemize}
  \item if $C= V_i , C \in \Vp$, $x' \in C^{\M_\Gamma}$ by the definition of $\M_\Gamma$.
 \item if $C$ is a classical formula, the proof is standard.
      \item if $C = (A \clsr B)$: since $\Gamma$ is saturated, for every $y \in \mathrm{Lab}_\Gamma$ we have either $x:\neg \Box\neg A \in \Gamma $ and $y:\neg B \in \Gamma$, or $y:B\in \Gamma$ and $y:\neg\Box_x\neg A \in \Gamma$. By \IV, in the first case we get $A^{\M_\Gamma} \neq \emptyset$ and $y \notin B^{\M_\Gamma}$, and in the second we have that $y \in (\neg\square_x\neg A)^{\M_\Gamma}$ which also entails $A^{\M_\Gamma} \neq \emptyset$. Thus, by claim \ref{claim:clsrsemantics}, we have $x \in (A \clsr B)^{\M_\Gamma}$.

    \item if $C = \neg(A \clsr B)$: by the saturation conditions, we have 3 cases.
    
    (a) $x:\square\neg A \in \Gamma$. By application of the rule $(T\square)$, for all label $y$, $y:\neg A \in \Gamma$. By our induction hypothesis, $A^{\M_\Gamma} = \emptyset$, and so $x \in (\neg(A \clsr B))^{\M_\Gamma}$.
    
    (b) $x:B \in \Gamma$. By \IV, $x \in B^{\M_\Gamma}$, and so $x \in (\neg(A \clsr B))^{\M_\Gamma}$ by axiom (4).
    
    (c) $x:\neg A, x:\neg B$ are in $\Gamma$, and there is a label $y$ such that $y:B, y:\square_x \neg A$ are in $\Gamma$. By induction hypothesis, we have $y\in B^{\M_\Gamma}$ and $y \in (\square_x \neg A)^{\M_\Gamma}$, so that by claim \ref{claim:clsrsemantics}, we have $x \in (\neg(A \clsr B))^{\M_\Gamma}$

 \item if $C = \Box_y \neg A$, by saturation we have: for all $z$, if $z <_ x \in \Gamma$ then $z:\neg A \in \Gamma$ and  $z:\Box_y \neg A \in \Gamma$. Then by \IV, we have that for all $z$, if  $z \prec_y x$ then $z \in  (\neg A)^{\M_\Gamma}$ which means that  $x \in  (\Box_y \neg A)^{\M_\Gamma}$ 
 
 \item    if $C =  \neg \Box_y \neg A$, by saturation we have either $y: A \in \Gamma $ or $y:\neg A \in \Gamma$. In the first case, since $\prec_y$ satisfies centering we have $y \prec_y x$, and by \IV, $y \in A^{\M_\Gamma}$. Thus $x \in (\neg\square_y\neg A)^{\M_\Gamma}$.
 In the second case, by saturation there is $z \in \mathrm{Lab}_\Gamma$ such that $z<_y x \in \Gamma$, $z: A \in \Gamma$. By \IV and the definition of $\prec_y$, we conclude that  $x \in  (\neg \Box_y\neg A)^{\M_\Gamma}$.     
  \end{itemize}
\end{proof}


\section{Termination of the Tableau Calculus}

The calculus presented above can lead to  non-terminating computations due to the interplay between the rules  which generate new labels (the dynamic rules $ (F2\clsr)$, $(F\square)$ and $(F\square_x)$) and the static rule  $ (T\clsr)$   which generates formula    $\neg \square_x A$ to which   $(F\square_x)$ may again be applied. 
Our calculus can be made terminating by defining a systematic procedure for applying the rules and by introducing appropriate blocking conditions. The systematic procedure simply prescribes to apply static rules as far as possible before applying dynamic rules. To prevent the generation of an infinite tableau set, we put some restrictions on the rule's applications. The restrictions on all rules except $(F2\clsr)$ and $(F2\square_x)$ are easy and prevent redundant applications of the rules.  We call the restrictions on $(F2\clsr)$ and $(F2\square_x)$ blocking conditions in analogy with standard conditions for getting termination in modal and description logics tableaux; they prevent the generation of infinitely many labels by performing a kind of loop-checking.

To this aim, we first define a total  ordering $\sqsubset$ on the labels of a tableau set such that
 $x \sqsubset y$ for all labels $x$ that are already in the tableau when $y$ is introduced. If $x \sqsubset y$, we will say that $x$ is older than $y$.

We define $\mathrm{Box}^{+}_{\Gamma,x,y}$ as the set of positive boxed formulas indexed by $x$ labelled by $y$ which are in $\Gamma$:
$\mathrm{Box}^{+}_{\Gamma,x,y} = \{ \square_x \neg A ~|~ y:\square_x \neg A \in \Gamma \}$
and $\Pi_\Gamma(x)$ as the set of non boxed formulas labelled by $x$: 
$\Pi_\Gamma(x) = \{ A ~|~ A \in \Lcsl \mbox{ and } x:A \in \Gamma \}$.

\begin{definition} \label{def:blocking}

  \noindent \emph{(Static and dynamic rules)} We call \emph{dynamic} the following rules: $(F2\clsr)$, $(F2\square_x)$ and $(F\square)$. We call \emph{static} all the other rules.

  \item \noindent \emph{(Rules restrictions)} 
  \begin{enumerate}
  \item Do not apply a static rule to $\Gamma$ if at least one of the consequences is already in it.
  \item Do not apply the rule $(F2\clsr)$ to a $x:\neg(A\clsr B), x:\neg A, x:\neg B$ 
    \begin{enumerate}
          \item if there exists some label $y$ in $\Gamma$ such that $y:B$ and $y:\square_x \neg A$ are in $\Gamma$.
          
          \item if there exists some label $u$ such that $u \sqsubset x$ and $\Pi_\Gamma(x) \subseteq \Pi_\Gamma(u)$.
    \end{enumerate}
  
  \item Do not apply the rule $(F2\square_x)$ to a $z:\neg \square_x \neg A, x:\neg A$
    \begin{enumerate}
          \item if there exists some label $y$ in $\Gamma$ such that $y<_x z$, $y: A$ and $y:\square_x \neg A$ are in $\Gamma$.
          
          \item if there exists some label $u$ in $\Gamma$ such that $u \sqsubset x$ and $\Pi_\Gamma(x) \subseteq \Pi_\Gamma(u)$.
          
          \item if there exists some label $v$ in $\Gamma$ such that $v\sqsubset z$ and $v:\neg\square_x\neg A \in \Gamma$ and $\mathrm{Box}^{+}_{\Gamma,x,z} \subseteq \mathrm{Box}^{+}_{\Gamma,x,v}$.
    \end{enumerate}
  
    \item Do not apply the rule $(F\square)$ to a $x:\neg\square\neg A$ in $\Gamma$ if there exists some label $y$ such that $y:A$ is in $\Gamma$.
  \end{enumerate}

  \noindent \emph{(Systematic procedure)}
    (1) Apply static rules as far as possible.
    (2) Apply a (non blocked) dynamic rule to some formula labelled $x$ only if no dynamic rule is applicable to a formula labelled $y$, such that $y\sqsubset x$.
\end{definition}

We prove that a tableau initialized with a  $\CSL$-formula always terminates provided it is expanded according to Definition \ref{def:blocking}.

\begin{theorem} Let $\Gamma$ be obtained from $\{x:A\}$, where $A$ is a $\CSL$-formula, by applying an arbitrary sequence of rules respecting definition \ref{def:blocking}. Then $\Gamma$ is finite.
\end{theorem}

\begin{proof}
 Suppose by absurdity that $\Gamma$ is not finite. Since the static rules (and also the $(F\square)$ rule) may  only add a finite of number of formulas for each label, $\Gamma$ must contain an infinite number of labels generated by the dynamic rules, either (F2$\clsr$) or (F2$\Box$) (or both).

Let $\Gamma$ contain infinitely many labels introduced by (F2$\clsr$).    Since the number of negative $\clsr$ formulas is finite, there must be one formula, say $\lnot (B \clsr C)$, such that for an infinite sequence of labels  $x_1, \ldots, x_i, \ldots$, $x_i: \lnot (B \clsr C) \in \Gamma$. By blocking condition (2b) we  then have that for every $i$, $\Pi_\Gamma(x_i) \not \subseteq 
\Pi_\Gamma(x_1), \ldots, \Pi_\Gamma(x_i) \not\subseteq \Pi_\Gamma(x_{i-1}) $. But this is impossible since each $\Pi_\Gamma(x_i)$ is finite (namely  bounded by 
$O(|A|)$) and the rules are non-decreasing wrt. $\Pi_\Gamma(x_i)$ (an application of a rule can never remove formulas from $\Pi_\Gamma(x_i)$).

Let now  $\Gamma$ contain infinitely many labels introduced by (F2$\Box$). That is to say, $\Gamma$ contains $x_i: \lnot \Box_{y_i} \lnot B$ for infinitely many $x_i$ and $y_i$. If all $y_i$ are distinct,  $\Gamma$ must contain in particular infinitely many formulas $x: \lnot \Box_{y_i}$ for a fixed $x$. The reason is that $x_i: \lnot \Box_{y_i}\lnot B$ may only be introduced by applying (T$\clsr$), thus there must be  infinitely many $y_i: B \clsr C  \in \Gamma$. By the systematic procedure, the rule (T $\clsr$) has been applied to a label $x$ for every $y_i: B \clsr C  \in \Gamma$ generating  $x: \lnot \Box_{y_i}\lnot B$ for all $y_i$. But then we can find a contradiction with respect to blocking condition (3b) as in the previous case, since for each $i$ we would have $\Pi_\Gamma(y_i) \not \subseteq \Pi_\Gamma(y_1), \ldots, \Pi_\Gamma(y_i) \not\subseteq \Pi_\Gamma(y_{i-1}) $ . We can conclude that $\Gamma$ cannot contain $x_i: \lnot \Box_{y_i} \lnot B$, for infinitely many distinct $y_i$ and distinct $x_i$. We are left with the case $\Gamma$ contains $x_i: \lnot \Box_{y} \lnot B$ for a fixed $y$ and infinitely many $x_i$. In this case, by blocking condition (3c), we have that for each $i$, $\mathrm{Box}^+_{\Gamma,y, x_i} \not \subseteq \mathrm{Box}^+_{\Gamma,y, x_1}, \ldots, \mathrm{Box}^+_{\Gamma,y, x_i} \not\subseteq \mathrm{Box}^+_{\Gamma,x_{i-1}}$. But again this is impossible given the fact that each $\mathrm{Box}^+_{\Gamma,y, x_i}$ is finite (bounded by $O(|A|)$) and that the rules are non-decreasing wrt. the sets $\mathrm{Box}^{+}_\Gamma$.
\end{proof}

To prove completeness, we will consider tableau sets saturated under blocking. A tableau set $\Gamma$ is saturated under blocking iff (a) it is build according to Definition \ref{def:blocking} (b) No further rules can be applied to it. It is easy to see that if $\Gamma$ is saturated under blocking, it satisfies all the saturation conditions in Definition \ref{def:saturation} except possibly for conditions $(F\clsr)$.(iii) and $(F\square_x)$.(ii).

By the termination theorem, we get that any tableau set generated from an initial set containing just a $\CSL$ formula, will be  either closed or saturated under blocking in a finite number of steps.

We now show that an open tableau set saturated under blocking can be extended to an open saturated tableau set,  that is satisfying all conditions of definition \ref{def:saturation}. By means of theorem \ref{th:tabsoundcomplete} we obtain the completeness of the terminating procedure.

\begin{theorem}
  If $\Gamma$ is saturated and open under blocking, then there exists an open and saturated set $\Gamma^*$ such that for all $A \in \Lcsl$, if $x:A \in \Gamma$ then $A \in \Gamma^*$.
\end{theorem}

Let $\Gamma$ be an open and saturated set under blocking. We will construct the set $\Gamma^*$ from $\Gamma$ in three steps. First, we consider formulas $z:\neg\square_x\neg A$ which are blocked by condition 3c (and not by 3b). We construct a set $\Gamma_1$ from $\Gamma$ which satisfies the saturation condition $(F\square_x)$ wrt. these formulas.

\begin{step}
  For each formula $z:\neg\square_x\neg A \in\Gamma$ for which condition $(F\square_x)$ is not fulfilled and that is blocked only by condition 3c, we consider the oldest label $u$ that blocks the formula. Therefore, the formula $u:\neg\square_x\neg A$ is in $\Gamma$ and it is not blocked by condition 3c \footnote{If it was, let $v$ older than $u$ the label which causes the blocking. Then $v$ will also block $u:\neg\square_x\neg A$, contradiction, as $u$ is by hypothesis the oldest label blocking $z:\neg\square_x\neg A$.}. Since $z:\neg\square_x\neg A$ is not blocked by condition 3b, $u:\neg\square_x\neg A$ is not blocked for this condition either, and thus the rule $(F2\square_x)$ has been applied to it. Hence there exists a label $y$ such that $y:A, y:\square_x \neg A$ and $y <_x u$ are in $\Gamma$. We then add $y <_x z$ to $\Gamma$.
  We call $\Gamma_1$ the resulting set.
\end{step}

\begin{claim2}\label{claim:gam1satopen}
  (I) $\Gamma_1$ is saturated, except for $(Mod)$ and the formulas $x:\neg(A\clsr B)$ and $z:\neg\square_x\neg A$ respectively blocked by condition 2b and 3b. (II) It is open.
\end{claim2}

The step 2 will now build a set $\Gamma_2$ saturated wrt. $(Mod)$ from $\Gamma_1$.

\begin{step}
  For each $y <_x z \in \Gamma$, if $\mathrm{Box}^{+}_{\Gamma,x,z} \subset \mathrm{Box}^{+}_{\Gamma,x,y}$, then for each $z_0$ such that $\mathrm{Box}^{+}_{\Gamma,x,z_0} = \mathrm{Box}^{+}_{\Gamma,x,z}$ we add $y <_x z_0 $ to $\Gamma_1$.
  We call $\Gamma_2$ the resulting set.
\end{step}

\begin{claim2}\label{claim:gam2satopen}
  (I) $\Gamma_2$ is saturated except for the formulas $x:\neg(A\clsr B)$ and $z:\neg\square_x\neg A$ respectively blocked by condition 2b and 3b. (II) It is open.
\end{claim2}

We will now consider the formulas blocked by conditions 2b and 3b, and finally build a set $\Gamma_3$ saturated wrt. all rules from $\Gamma_2$.

\begin{step}
For each label $x$ such that there is  a formula $x:\neg(A \clsr B) \in \Gamma$ or $z:\neg\square_x\neg A \in \Gamma$ respectively blocked by condition 2b or 3b, we let $u$ be the oldest label which caused the blocking. We then construct the set $\Gamma_3$ by the following procedure:
\begin{enumerate}
    \item we remove from $\Gamma_2$ each relation $<_x$, and all formulas $v:\neg\square_x\neg A$ and $v:\square_x\neg A$ ($v \in \mathrm{Lab}_\Gamma$).
    \item For all label $z \in \mathrm{Lab}_\Gamma$ such that $z\neq x$, we add $x <_x z$.
    \item For all labels $z,v \in \mathrm{Lab}_\Gamma$ such that $z \neq x$, if $v <_u z \in \Gamma_2$, then we add $v <_x z$.
    \item For each $v:\square_u\neg A \in \Gamma$, if $A \in \Pi_\Gamma(x)$ we then add $v:\square_x\neg A$.
    \item For each $v:\neg\square_u\neg A \in \Gamma$ such that $v \neq x$, we add $v:\neg\square_u\neg A$ 
    \item For each formula $A \in \Pi_\Gamma(x)$, we add $x:\square_x A$.
  \end{enumerate}
\end{step}

\begin{claim2}\label{claim:gam3satopen}
  (I) $\Gamma_3$ is saturated wrt. all rules. (II) It is open.
\end{claim2}

We then let $\Gamma^* = \Gamma_3$. It is easy to see that for all formulas $A \in \Lcsl$, if $x:A \in \Gamma$ then $x:A \in \Gamma^*$, as none of these formulas are removed by the construction of $\Gamma^*$.

\longreport{}{

We now prove the precedent claims.
\begin{claim2} \label{claim:misc}
\begin{enumerate}
  \item If $y <_x z \in \Gamma$, then $\mathrm{Box}^{+}_{\Gamma,x,z} \subseteq \mathrm{Box}^{+}_{\Gamma,x,y}$.
  
  \item If $y <_x z$ is in $\Gamma_1$, then $\mathrm{Box}^{+}_{\Gamma,x,z} \subseteq \mathrm{Box}^{+}_{\Gamma,x,y}$.
  
  \item If $z:\neg\square_x\neg A$ is blocked by condition 3c and if $u$ is the oldest label (according to $\sqsubset$) blocking it, then $u:\neg\square_x\neg A$ is not blocked by condition 3c.
  
  \item If $z:\neg\square_x\neg A$ and $x:\neg( A \clsr B)$ are blocked by condition 3b or  2b, and if $u$ is the oldest label blocking it, then $u$ cannot be blocked by condition 3b nor 2b.
\end{enumerate}
\end{claim2}

\begin{proof}
  \begin{enumerate}
    \item Trivial, since $\Gamma$ is saturated wrt. $(T\square_x)$.
    
    \item If $y <_x z \in \Gamma$, we are in the precedent case. If not, then $y <_x z$ was added by step 1. Thus there is a formula $z:\neg\square_x\neg A$ blocked by condition 3c. Let $u$ be the oldest label blocking it. We then have, by definition of blocking condition 3c, $\mathrm{Box}^{+}_{\Gamma,x,z} \subseteq \mathrm{Box}^{+}_{\Gamma,x,u}$.  We also have, by definition of step 1, that $y <_x u \in \Gamma$. Thus, as shown in part 1 of this lemma, we have $\mathrm{Box}^{+}_{\Gamma,x,u} \subseteq \mathrm{Box}^{+}_{\Gamma,x,y}$. We can now conclude that $\mathrm{Box}^{+}_{\Gamma,x,z} \subseteq \mathrm{Box}^{+}_{\Gamma,x,y}$.
    
    \item Suppose that $u$ is the oldest label blocking $z:\neg\square_x\neg A$, and that $u:\neg\square_x\neg A$ is blocked by $v$. By definition of the blocking condition, we have that $v \sqsubset u \sqsubset z$, and $\mathrm{Box}^{+}_{\Gamma,x,z} \subseteq \mathrm{Box}^{+}_{\Gamma,x,u} \subseteq \mathrm{Box}^{+}_{\Gamma,x,v}$. Then we have that $v$ also blocks $z:\neg\square_x\neg A$: contradiction, as $u$ should be, by hypothesis, the oldest label blocking this formula.
    
    \item Suppose that $u$ blocks $z:\neg\square_x\neg A$ or $x:\neg(A \clsr B)$ by condition 3b or 2b. Then we have $u \sqsubset x$ and $\Pi(x) \subseteq \Pi(u)$. Now suppose that there is a formula $u:\neg(A \clsr B)$ or $w:\neg\square_u\neg A$ ($w \in \mathrm{Lab}_\Gamma)$ blocked by condition 2b or 3b by a label $v$. Then $v \sqsubset u$ and $\Pi(u) \subseteq \Pi(v)$. Since we have $v \sqsubset x$ and $\Pi(x) \subseteq \Pi(v)$, we would have that $v$ also blocks $x$: contradiction, as $u$ is by hypothesis the oldest label blocking it.
  \end{enumerate}

\end{proof}

\begin{proof}[Claim \ref{claim:gam1satopen}-(I)]
  As $\Gamma$ is saturated under blocking, and $\Gamma \subseteq \Gamma_1$, we only have to check that the formulas $z:\neg\square_x\neg A$ which were blocked by condition 3c (and not by 3b) satisfy the saturation condition $(F\square_x)$. As we add a preferential relation $y <_x z$, we also need to check the saturation wrt. to $(T\square_x)$.
  \begin{itemize}
    \item $(F2\square_x)$: Let $z:\neg\square_x\neg A$ blocked by condition 3b (and only by this condition). By construction, there is a label $y$ such that $y:A, y:\square\neg A$ and $y <_x z$ are in $\Gamma_1$, so the formula $z:\neg\square_x\neg A$ satisfies the saturation condition.
  
    \item $(T\square_x)$: Let $z:\square_x \neg C$ and $v <_x z$ in $\Gamma_1$. We have two cases: (1) $v <_x z$ is already in $\Gamma$, and as $\Gamma$ is saturated wrt. $(T\square_x)$, the saturation condition holds in $\Gamma_1$. (2) $v <_x z$ was not in $\Gamma$, and so that $v <_x z$ was added by construction of $\Gamma_1$. By claim \ref{claim:misc}-2, we obtain $\mathrm{Box}^{+}_{\Gamma,x,z} \subseteq \mathrm{Box}^{+}_{\Gamma,x,v}$ and  the proof is trivial.
  \end{itemize}
\end{proof}

\begin{proof}[Claim \ref{claim:gam1satopen}-(II)]
  Since $\Gamma_1$ is obtained by adding only preferential formulas to $\Gamma$, and $\Gamma$ is open, we only have to check closure condition (ii).
    
    Suppose that $\Gamma_1$ is closed by definition \ref{def:closed-branch}-(ii): $y <_x z$ and $z <_x y$ are in $\Gamma_1$. Then we have 3 cases:
    (1) $y <_x z$ and $z <_x y$ are in $\Gamma$: contradiction, $\Gamma$ is open.
    (2) $z <_x y$ is in $\Gamma$, but $y <_x z$ is not. Therefore $y <_x z$ has been added by construction of $\Gamma_1$, and so by claim \ref{claim:misc}-2 we have $\mathrm{Box}^{+}_{\Gamma,x,z} \subseteq \mathrm{Box}^{+}_{\Gamma,x,y}$. Since  $z <_x y$ is in $\Gamma$ we have by claim \ref{claim:misc}-1 $\mathrm{Box}^{+}_{\Gamma,x,y} \subseteq \mathrm{Box}^{+}_{\Gamma,x,z}$ (*). Thus we have: $\mathrm{Box}^{+}_{\Gamma,x,z} = \mathrm{Box}^{+}_{\Gamma,x,y}$. By definition of $\Gamma_1$, we also have that $y:\square_x\neg A$ is in $\Gamma$. By the inclusion (*), we have that $z:\square_x\neg A$ must also be in $\Gamma$. But $z:\neg\square_x\neg A \in \Gamma$: contradiction, $\Gamma$ is open. The case where $y <_x z$ is in $\Gamma$ but $z <_x y$ is not is symmetric.
    (3) Neither $y <_x z$ nor $z <_x y$ are in $\Gamma$. Both formulas has been added by construction of $\Gamma_1$. Thus there are some $y:\neg\square_x\neg A$ and $z:\neg\square_x\neg B$ in $\Gamma$ blocked by condition 3c. By construction of $\Gamma_1$, if $y <_x z$ and $z<_x y$ were added, then we must have $y:\square_x \neg B$ and $z:\square_x \neg A$ in $\Gamma$. Using claim \ref{claim:misc}-2, if $y <_x z$ and $z <_x y$ are in $\Gamma_1$, then we must have $\mathrm{Box}^{+}_{\Gamma,x,z} \subseteq \mathrm{Box}^{+}_{\Gamma,x,y}$ and $\mathrm{Box}^{+}_{\Gamma,x,y} \subseteq \mathrm{Box}^{+}_{\Gamma,x,z}$. Thus $\mathrm{Box}^{+}_{\Gamma,x,z} = \mathrm{Box}^{+}_{\Gamma,x,y}$. Thus we have $z:\square_x \neg B$ and $y:\square_x \neg A$ also in $\Gamma$: contradiction, $\Gamma$ is open.

\end{proof}

\begin{proof}[Claim \ref{claim:gam2satopen}-(I)]
  We have to check saturation wrt. $(Mod)$ and $(T\square_x)$. With regards to $(Mod)$, let $u <_x z \in \Gamma_2$. We have several cases:
  (1) $y <_x z \in \Gamma$: trivial, $\Gamma$ being saturated wrt. $(Mod)$.
  
  (2) $y <_x z \in \Gamma_1$ but not in $\Gamma$, $y <_x z$ must have been added by construction of $\Gamma_1$. Let $u \in \mathrm{Lab}_\Gamma$, we have two cases: (2a) either $u <_x z$ or $y <_x u$ are in $\Gamma_1$: the saturation condition is then satisfied.
  (2b) neither $u <_x z$ nor $y <_x u$ are in $\Gamma_1$. By construction of $\Gamma_1$, we have that there is $z:\neg\square_x\neg A \in \Gamma$ which is blocked by $v:\neg\square_x\neg A$ by condition 2b, and we have that $y:\square_x \neg A$ and $y <_x v$ are in $\Gamma$. We also have $\mathrm{Box}^{+}_{\Gamma,x,z} \subseteq \mathrm{Box}^{+}_{\Gamma,x,y}$ (claim \ref{claim:misc}-2). Since $\square_x \neg A \in \mathrm{Box}^{+}_{\Gamma,x,y}$ but not in $\mathrm{Box}^{+}_{\Gamma,x,z}$ ($\Gamma_1$ would be closed), we have that  $\mathrm{Box}^{+}_{\Gamma,x,z} \subset \mathrm{Box}^{+}_{\Gamma,x,y}$. As $y <_x v$ is in $\Gamma$, $(Mod)$ has been applied to it with $u$ so either $y <_x u$ or $u <_x v$ are in $\Gamma$. The first case cannot occur by our hypothesis, so we have that $u <_x v$ is in $\Gamma$. $(Mod)$ had also been applied to it with $z$, so either $u <_x z$ is in $\Gamma$, or $z <_x v$ is in $\Gamma$. The first case being not possible by hypothesis, we have $z <_x v \in \Gamma$, and so $\mathrm{Box}^{+}_{\Gamma,x,v} \subseteq \mathrm{Box}^{+}_{\Gamma,x,z}$ (by claim \ref{claim:misc}-1). As $v$ blocks $z$ by condition 3c, we also have $\mathrm{Box}^{+}_{\Gamma,x,z} \subseteq \mathrm{Box}^{+}_{\Gamma,x,v}$ (by definition of blocking condition). So $\mathrm{Box}^{+}_{\Gamma,x,v} = \mathrm{Box}^{+}_{\Gamma,x,z}$, and by definition of $\Gamma_2$, as $u <_x v \in \Gamma$, $u <_x z$ is in $\Gamma_2$.
  
  (3) $y <_x z \in \Gamma_2$ but not in $\Gamma_1$. So $y <_x z$ has been added by construction of $\Gamma_2$. We then have that there is some $y <_x v$ in $\Gamma$ such that $\mathrm{Box}^{+}_{\Gamma,x,v} = \mathrm{Box}^{+}_{\Gamma,x,z}$. Then for all $u \in \mathrm{Lab}_\Gamma$, we have two cases:
  (3a) Either $u <_x z$ or $y <_x u$ are in $\Gamma_1$: this case is easy.
  (3b) Neither $u <_x z$ nor $y <_x u$ are in $\Gamma_1$. As $y <_x v \in \Gamma$, $(Mod)$ had been applied to it with $u$: so either $y <_x u$ or $u <_x v$ are in $\Gamma$. The first case is impossible by hypothesis, so $u <_x v$ is in $\Gamma$. As $\mathrm{Box}^{+}_{\Gamma,x,v} = \mathrm{Box}^{+}_{\Gamma,x,z}$ and by construction of $\Gamma_2$, $u <_x z$ is then in $\Gamma_2$.
  
  As step 2 add some preferential relations $y <_x z$, we have to check the saturation wrt. $(T\square_x)$. By definition of Step 2, if $y <_x z$ was added in $\Gamma_2$, we have $\mathrm{Box}^{+}_{\Gamma,x,z} \subseteq \mathrm{Box}^{+}_{\Gamma,x,y}$, and thus the saturation condition easily follows.
\end{proof}

\begin{proof}[Claim \ref{claim:gam2satopen}-(II)]
  The case of the closures conditions (i) and (iii) are trivial (as $\Gamma_1$ is open, and as step 2 only adds preferential formulas). We now consider the case of the closure condition (ii).
  
  Suppose that $y <_x z$ and $z <_x y$ are in $\Gamma_2$. Then we have several cases:
  
  (1) both formulas are in $\Gamma_1$: contradiction with the fact that $\Gamma_1$ is open.
  
  (2) $z <_x y \in \Gamma_1$ but $y <_x z$ is not. Then $z <_x y$ have been added by construction of $\Gamma_2$. So there is a label $v$ such that $y <_x v \in \Gamma$, $\mathrm{Box}^{+}_{\Gamma,x,v} = \mathrm{Box}^{+}_{\Gamma,x,z}$, and $\mathrm{Box}^{+}_{\Gamma,x,v} \subset \mathrm{Box}^{+}_{\Gamma,x,y}$. As $z <_x y \in \Gamma_1$, we also have that $\mathrm{Box}^{+}_{\Gamma,x,y} \subseteq \mathrm{Box}^{+}_{\Gamma,x,z}$ (claim claim \ref{claim:misc}-2), and so $\mathrm{Box}^{+}_{\Gamma,x,y} \subseteq \mathrm{Box}^{+}_{\Gamma,x,v}$, which leads to a contradiction.
  
  (3) neither $z <_x y$ nor $y <_x z$ are in $\Gamma_1$. Both formulas have been added by construction of $\Gamma_2$. So there are some labels $v$ and $w$ such that $z <_x v$ and $y <_x w$ are in $\Gamma$. Furthermore, we have $\mathrm{Box}^{+}_{\Gamma,x,v} \subset \mathrm{Box}^{+}_{\Gamma,x,z}$, $\mathrm{Box}^{+}_{\Gamma,x,w} \subset \mathrm{Box}^{+}_{\Gamma,x,y}$, and $\mathrm{Box}^{+}_{\Gamma,x,v} = \mathrm{Box}^{+}_{\Gamma,x,y}$ $\mathrm{Box}^{+}_{\Gamma,x,w} = \mathrm{Box}^{+}_{\Gamma,x,z}$. So we can conclude that $\mathrm{Box}^{+}_{\Gamma,x,y} \subset \mathrm{Box}^{+}_{\Gamma,x,z}$ and $\mathrm{Box}^{+}_{\Gamma,x,z} \subset \mathrm{Box}^{+}_{\Gamma,x,y}$: we get a contradiction.
\end{proof}

\begin{proof}[Claim \ref{claim:gam3satopen}-(I)]
\begin{itemize}
  \item $(T\sqcap)$, $(N\sqcap)$ and $(NEG)$: trivial, as $\Gamma_2$ was saturated with respect to those rules, and considering the fact that for all $A \in \Lcsl$, if $x:A \in \Gamma_2$ then $x:A \in \Gamma_3$.
  
  \item $(F1\clsr)$: trivial, as $\Gamma_2$ is saturated with respect to this rule and the formulas added by it are not removed in the construction of $\Gamma_3$.
  
  \item $(T\clsr)$: if $x:(A \clsr B)$ is in $\Gamma_3$, then it must be in $\Gamma_2$. If $x$ is not blocked, it's easy, as $\Gamma_2$ is saturated wrt. this rule, so either $x:\neg\square\neg A,~y:\neg B$ or $y:B, y:\neg\square_x \neg A$ must be in $\Gamma_3$. 
  
  Otherwise, if $x$ is blocked by condition 2b or 3b, let $z$ be the oldest label blocking it. For each label $y$ in $\Gamma_3$ we have two cases: either $y:\neg B \in \Gamma_3$ or $y:B \in \Gamma_3$. The first case is easy, $y:\neg B$ must have been in $\Gamma_2$, and by saturation, $y:\neg\square\neg A$ too. As this formula cannot be removed between $\Gamma_2$ and $\Gamma_3$, we have the saturation. In the second case, $y:B$ must have been in $\Gamma_2$. We have, as $x$ is blocked by $z$, $\Pi_\Gamma(x) \subseteq \Pi_\Gamma(z)$, and so $z:(A\clsr B)$ is in $\Gamma_2$. Since $\Gamma_2$ is saturated wrt. $(T\clsr)$ and $y:B \in \Gamma_2$, and as $z$ is not blocked by condition 2b or 3b (by claim \ref{claim:misc}-4), we have $y:\neg\square_z\neg A \in \Gamma_2$. By definition of $\Gamma_3$, we then have $y:\neg\square_x\neg A \in \Gamma_3$. So $\Gamma_3$ is saturated wrt. to $(T\clsr)$.
  
  \item $(F2\clsr)$: Let $x$ be blocked by condition 2b or 3b (the case where $x$ is non blocked is trivial), and let $z$ be the oldest label blocking it. As $\Pi_\Gamma(x) \subseteq \Pi_\Gamma(z)$, $z:\neg(A \clsr B), z:\neg A, z:\neg B$ must be in $\Gamma_2$. As $z$ is not blocked (by claim \ref{claim:misc}-4), $(F2\clsr)$ must have been applied to it. So there exists a label $u$ such that $u:B, u:\square_z \neg A$ are in $\Gamma_2$. By construction of $\Gamma_3$, we have that $u:\square_x \neg A$ is in $\Gamma_3$, making it saturated wrt. $(F2\clsr)$.
  
  \item $(F1\square_x)$: if $x$ is not blocked by condition 2b or 3b, it is trivial. Otherwise, let $v$ be the oldest label blocking $x$. As $z:\neg\square_x \neg A$ is in $\Gamma_3$, $z:\neg\square_v \neg A$ must be in $\Gamma_2$ (by construction of $\Gamma_3$). As $v$ is not blocked (by claim \ref{claim:misc}-4), $(F1\square_x)$ must have been applied to $z:\neg\square_v \neg A$ from which we obtain the conclusion.

  \item $(T\square_x)$: if $x$ is not blocked by 2b or 3b, the proof is easy. Otherwise, let $v$ be the oldest label blocking it. As $z:\square_x \neg A$ and $y <_x z$ are in $\Gamma_3$, $z:\square_v \neg A$ and $y <_v z$ are in $\Gamma_2$ (and note that $x:\neg A$ must be in $\Gamma_2$ too). As $v$ is not blocked (by claim \ref{claim:misc}-4), the rule $(T\square_x)$ have been applied to these formulas, and so $y:\neg A$ and $y:\square_v \neg A$ are in $\Gamma_2$, and so in $\Gamma_3$.
  
  \item $(F2\square_x)$: if $x$ is not blocked by conditions 2b or 3b, the proof is easy. Otherwise, let $v$ be the oldest label blocking it. As $z:\neg \square_x \neg A$ and $x:\neg A$ are in $\Gamma_3$, $z:\neg\square_v\neg A$ and $x:\neg A$ must be in $\Gamma_2$. Moreover, as $v$ is not blocked (by claim \ref{claim:misc}-4), the rule $(F2\square_x)$ has been applied to these formulas. So there exists a label $u$ such that $u <_v z$, $u:A$ and $u:\square_v \neg A$ are in $\Gamma_2$. And so, by definition of $\Gamma_3$, $u <_x z$, $u:A$ and $u:\square_x \neg A$ are also in $\Gamma_3$.
  
  \item $(T\square)$ and $(F\square)$: trivial.
  
  \item $(Mod)$: if $x$ is not blocked by conditions 2b or 3b, the relation $<_x$ was already saturated in $\Gamma_2$, and not modified in $\Gamma_3$. If $x$ is blocked (by condition 2b or 3b), let $v$ be the oldest label blocking it. As $v$ is not blocked, the relation $<_v$ is saturated for $(Mod)$ in $\Gamma_2$. Let $z <_x u \in \Gamma_3$. Note that, by definition of $\Gamma_3$, $u \neq x$. If $z=x$, for all labels $y\neq x$ have $x <_x y$ by construction of $\gamma_3$. If $z \neq x$, then for all $y$ we have two cases: (a) $y=x$: then we have $x <_x u$ by construction of $\Gamma_3$. (b) $y\neq x$: then $z <_v u$ must have been in $\Gamma_2$. As $\Gamma_2$ is saturated and $v$ not blocked, either $y <_v u$ or $z <_v y$ are in $\Gamma_2$, and so either $y <_x u$ or $u <_x z$ are in $\Gamma_3$ by construction.
  
  By definition of step 3, if $y=x$ or $z=x$, at least one of these formula is not in $\Gamma_3$, contradicting our hypothesis.
  
  \item $(Cent)$: easy, either by saturation of $\Gamma_2$ if $x$ is not blocked by 2b or 3b, or by construction of $\Gamma_3$ in the other case.
\end{itemize}
\end{proof}

\begin{proof}[Claim \ref{claim:gam3satopen}-(II)]
None of the closure conditions could occur in $\Gamma_3$:
\begin{itemize}
  \item Suppose that $x:C$ and $x:\neg C$ are in $\Gamma_3$. If $C\in \Lcsl$, then $x:C$ and $x:\neg C$ must be in $\Gamma_2$: contradiction because $\Gamma_2$ is open.
  
  If $C = \square_z \neg A$, then two cases: \\
  (a) $z$ is not blocked by condition 2b or 3b. Then $x:\square_z \neg A$ and $z:\neg\square_z \neg A$ are in $\Gamma_2$ which leads to a contradiction as $\Gamma_2$ is open. \\ 
  (b) $z$ is blocked by condition 2b or 3b. Let $v$ be the oldest label blocking it. Then, by construction of $\Gamma_3$, $z:\square_v \neg A$ and $z:\neg\square_v\neg A$ are in $\Gamma_2$: contradiction as $\Gamma_2$ is open.
  
  \item Suppose that $y <_x z$ and $z <_x y$ are in $\Gamma_3$. If $x$ is not blocked by condition 2b or 3b, both formulas are in $\Gamma_2$, which leads to a contradiction. If $x$ is blocked by condition 2b or 3b, let $v$ be the oldest label blocking it. Suppose that $y \neq x \neq y$. Then, by construction of $\Gamma_3$, $z <_v y$ and $y <_v z$ are in $\Gamma_2$: contradiction. 
  
  \item Suppose that $x:\neg\square_x \neg A$. This formula cannot have been added by step 3 (by definition of this step), so it must have been in $\Gamma_2$ (and then $x$ must be not blocked by condition 2b or 3b): contradiction, as $\Gamma_2$ is open.
\end{itemize}
\end{proof}

}

The tableaux procedure described in this section gives a decision procedure for $\CSL$. To estimate its
complexity, let the length of $A$, the initial formula, be $n$.
It is not hard to see that any tableau set saturated under blocking may contain at most $O(2^n)$
labels. As matter of fact by the blocking conditions no more than $O(2^n)$ labels can be introduced by dynamic rules $F2\clsr$ and $F2\Box_x$. Thus
a saturated set under blocking will contain most $O(2^n)$ tableau formulas.
We can hence devise a non deterministic procedure that guesses an open tableau set in $O(2^n)$
steps. This shows that our tableau calculus gives a NEXPTIME decision procedure
for $\CSL$. In light of the results contained in \cite{Wolter:LPAR05} our procedure is not optimal, since it is shown that this logic is EXPTIME complete. We will study possible optimization (based for instance on caching techniques) in subsequent work.

\section{Conclusion}\label{section:conclusion}

In this paper, we have studied the logic $\CSL$ over minspaces, and we have obtained two main results: first we have provided  a direct, sound and complete axiomatisation of this logic. Furthermore, we have defined a tableau calculus, which gives a decision procedure for this logic.

In \cite{hudstadteETalsJelia06}, a tableau algorithm is proposed to handle logics for metric spaces  comprising distance quantifiers of
the form $\exists^{<a}A$ and alike, where $a$ is positive integer (together with an interior and a closure operator).
As observed in \cite{Wolter08}, the operator $\clsr$ can be defined in a related logic that allows  quantification
on the parameters in distance  quantifiers.
The methods proposed in \cite{hudstadteETalsJelia06} make use of an elegant relational translation to handle distance quantifiers with fixed parameters.
However, it is not clear if they  can be adapted to handle also the concept similarity operator.

There are a number of issues to explore in future research. The decision procedure  outlined in the previous section is not guaranteed to have an optimal complexity,  so that we can  consider how to improve our calculus in order to match this upper bound. Another issue is the extension of our results to symmetric minspaces, and possibly to other classes of models. Finally, since one original motivation of $\CSL$ is to reason about concept similarity in ontologies, and particularly in description logics, we plan to study further its integration with significant languages of this family.


\bibliographystyle{plain} 
\bibliography{ccs-article}

\end{document}